\documentclass[conference]{IEEEtran}
\IEEEoverridecommandlockouts

\usepackage{cite}
\usepackage{amsmath,amssymb,amsfonts}
\usepackage{algorithmic}
\usepackage{graphicx}
\usepackage{textcomp}
\usepackage{xcolor}

\usepackage{bbding}
\usepackage{booktabs}
\usepackage{multirow}
\usepackage{threeparttable}

\usepackage{makecell}

\usepackage{marvosym}

\def\BibTeX{{\rm B\kern-.05em{\sc i\kern-.025em b}\kern-.08em
    T\kern-.1667em\lower.7ex\hbox{E}\kern-.125emX}}
\begin{document}

\title{VAGU \& GtS: LLM-Based Benchmark and Framework for Joint Video Anomaly Grounding and Understanding 
}

\author{
	\IEEEauthorblockN{
		Shibo Gao\IEEEauthorrefmark{1}\IEEEauthorrefmark{2},
		Peipei Yang\IEEEauthorrefmark{2}\IEEEauthorrefmark{3},
		Yangyang Liu\IEEEauthorrefmark{2}\IEEEauthorrefmark{3},
		Yi Chen\IEEEauthorrefmark{3},
		Han Zhu\IEEEauthorrefmark{3},
		Xuyao Zhang\IEEEauthorrefmark{2}\IEEEauthorrefmark{3},
		Linlin Huang\IEEEauthorrefmark{1},
	}
	\IEEEauthorblockA{\IEEEauthorrefmark{1}Beijing Jiaotong University}
	\IEEEauthorblockA{\IEEEauthorrefmark{2}State Key Laboratory of Multimodal Artificial Intelligence Systems, Institute of Automation, Chinese Academy of Sciences}
	\IEEEauthorblockA{\IEEEauthorrefmark{3}School of Artificial Intelligence, University of Chinese Academy of Sciences}	
}

\maketitle

\begin{abstract}
	Video Anomaly Detection (VAD) aims to identify anomalous events in videos and accurately determine their time intervals. Current VAD methods mainly fall into two categories: traditional DNN-based approaches that focus on temporal localization, and LLM-based approaches that emphasize semantic understanding. Both anomaly understanding and grounding are essential for comprehensive video anomaly detection and can complement each other. However, no existing model or dataset supports both tasks simultaneously. To address this, we introduce VAGU (Video Anomaly Grounding and Understanding), the first benchmark to integrate both tasks. Each VAGU instance includes annotations for anomaly category, semantic explanation, precise temporal grounding and Video QA. We also provide multiple-choice Video QA for objective evaluation. Based on this dataset, we propose Glance then Scrutinize (GtS), a training-free framework guided by textual prompts.
	The framework first enables coarse localization of high-probability anomalous regions, followed by detailed anomaly interpretation and temporal boundary refinement.
	Additionally, we propose the JeAUG metric, which jointly evaluates semantic interpretability and temporal precision, overcoming the limitations of traditional metrics. Extensive experiments verify the effectiveness of our benchmark, framework, and evaluation metric.
\end{abstract}

\section{Introduction}

Video Anomaly Detection (VAD) aims to understand and temporally ground anomalous events in video sequences. In recent years, driven by the growing demand for real-time monitoring in industrial automation, intelligent surveillance, and smart transportation systems, VAD has emerged as a critical research frontier in computer vision and multimedia analytics~\cite{shanghaitech,learn-28-lu2020few,wacv_2022_fast,Intelligent_Automation_2022,ijcnn_ad_1_2022}.

\begin{figure}[h]
	\centering
	\includegraphics[width=\linewidth]{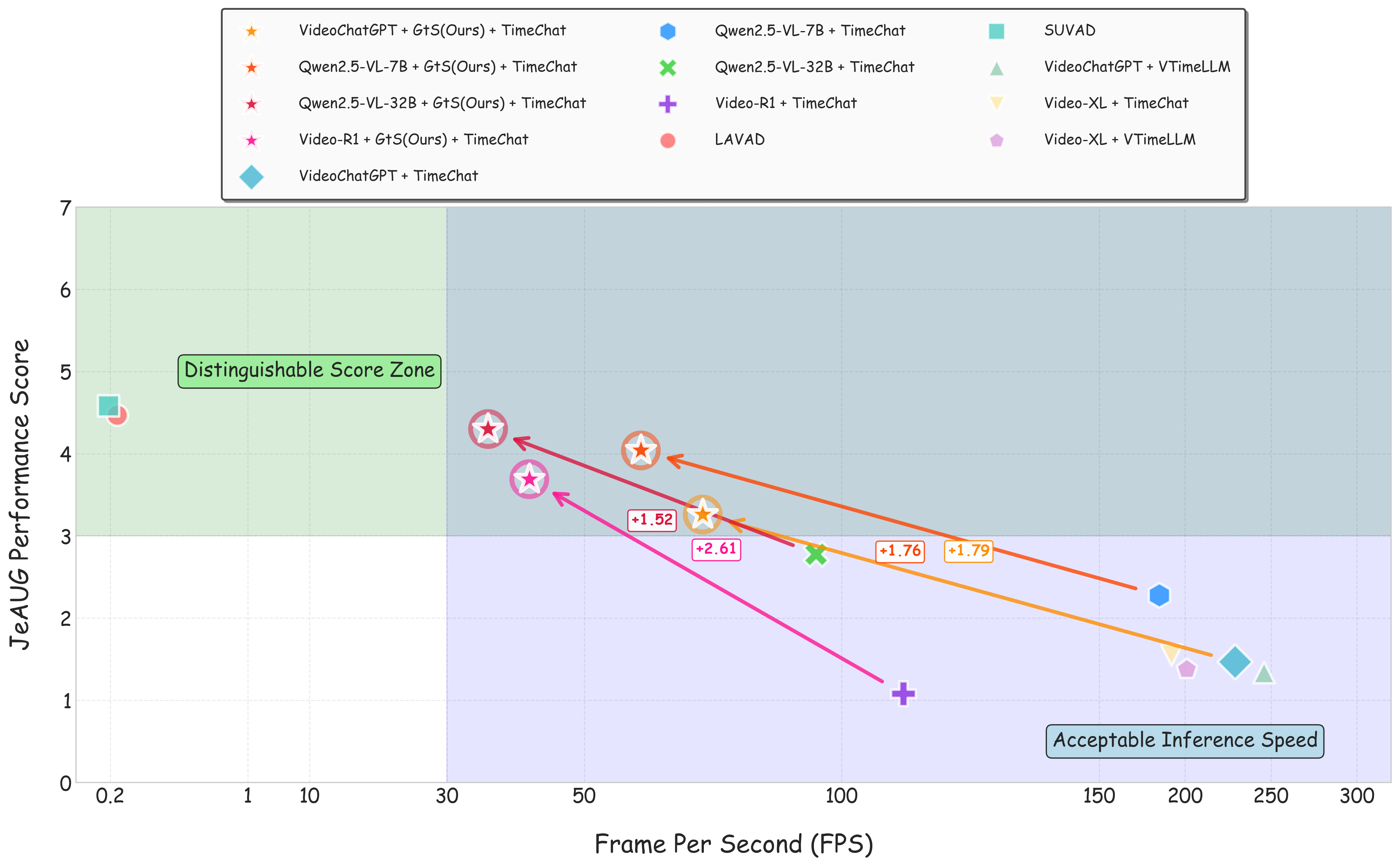}
	\caption{Comparison of our proposed framework with other LLM-based VAD methods in terms of inference speed and performance. It can be seen that our proposed framework achieves the best overall performance.}
	\label{fig:performance compare}
\end{figure}

The current video anomaly detection domain exhibits significant methodological bifurcation: traditional DNN-based methods (semi-supervised~\cite{ icdm_2022_great_again,liu2021hybrid,ECCV_2022_Jigsaw_Puzzles}, weakly-supervised~\cite{weakly_Dual_2023,weakly_Generative_2022,vadclip_tsa,vadclip_vadclip}, open-set VAD~\cite{open-set-ding2022catching,open-set-zhu2022towards,open-set-zhu2023anomaly} and else) and LLM-based methods respectively focus on temporal grounding and semantic understanding of anomalies, creating a "when-what" capability dissociation~\cite{HAWK,VERA,VANE-Bench}.
On one hand, traditional DNN-based methods learn normal/anomalous patterns in video sequences through video-level classification labels to capture anomalous events, yet they merely output temporal grounding results ("when anomalies occur") while lacking semantic understanding. On the other hand, emerging LLM-based methods can generate natural language descriptions leveraging LLMs' open-domain knowledge to answer "what anomalies occur", but universally neglect precise temporal grounding of anomaly onset/offset boundaries.

\begin{figure*}[h]
	\centering
	\includegraphics[width=\linewidth]{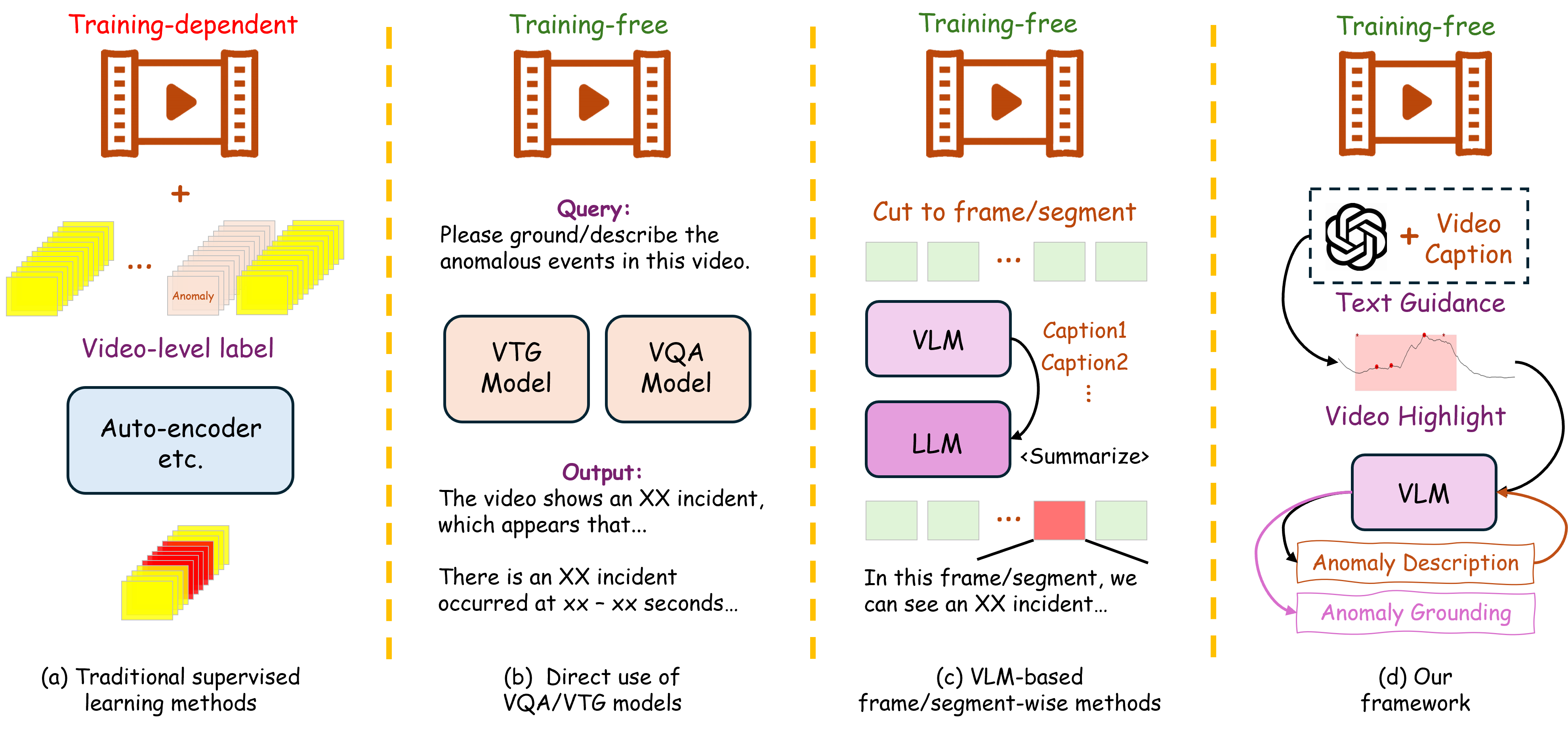}
	\caption{Comparison of our proposed framework with other VAD paradigms. Our proposed GtS framework can construct anomaly grounding and anomaly understanding capabilities based on existing multi-modal large language models under training-free conditions, while maintaining an acceptable inference speed.}
	\label{fig:differ}
\end{figure*}

Although some methods attempt joint grounding and understanding through vision-language models (VLMs) via frame/segment-wise video analysis~\cite{SUVAD,LAVAD,AnyAnomaly}, their prohibitively high computational overhead renders them unsuitable for meeting VAD's stringent real-time processing requirements.
Notably, due to the lack of a consistent standard for defining anomalous events, existing VQA/VTG models exhibit limited applicability when directly transferred to the VAD domain, as they face challenges with the openness of the problem~\cite{VAD-LLaMA}.

In light of these research, we systematically investigate three pivotal questions:

{\textbf{Q1:Why both importance of grounding \& understanding?}}

A1: VAD task necessitates simultaneous resolution of both temporal grounding ("when") and semantic comprehension ("what") queries~\cite{paper3,LAVAD,SUVAD,CUVA}. Additionally, these dual objectives exhibit intrinsic interdependence: temporal grounding provides critical contextual cues for semantic understanding, while semantic understanding feedback validates the plausibility of grounding results. Nevertheless, both DNN-based methods and LLM-based methods still fall short in establishing effective mechanisms for mutual assistance between these two dimensions.

{\textbf{Q2: Why do VQA/VTG models perform poorly when directly applied to VAD tasks?}}

A2: The fundamental reason lies in the absence of a consistent standard for defining the scope of anomalies. In VAD tasks, there are typically no explicit labels indicating specific anomalies within each video, and temporal grounding prompts are inherently missing. This makes it difficult for VQA models, which rely on predefined questions, to effectively handle open-ended anomaly-related queries such as "What anomalies are present in the video?" Similarly, VTG models depend on textual queries to ground relevant video segments, but in VAD, such queries are usually unavailable~\cite{Video-Llava,Video-XL,VideoChatGPT, VTimeLLM, mPLUG-Owl}.


{\textbf{Q3:Why training-free framework?}}

A3: Firstly, VAD tasks face data scarcity challenges: real-world normal/anomalous data exhibits extremely imbalanced distribution, with anomalous data collection difficulties and prohibitively high annotation costs. Furthermore, the collected datasets often fail to encompass all anomalous categories, which constrains the scalability of supervised learning methods\cite{paper3,open-vocabulary-detection-kim2023region}.
Secondly, real-world VAD applications demand exceptional generalizability, whereas supervised learning methods frequently exhibit degraded robustness in cross-scenario deployments.
Notably, although frame/segment-wise VLM-based methods can concurrently output grounding and comprehension, its computational overhead remains prohibitively high — processing a video of several tens of seconds may require tens of hours — thus fundamentally conflicting with VAD's real-time processing requirements.

Building upon these critical questions, we rethink the VAD task and propose the VAGU benchmark (Video Anomaly Grounding and Understanding) — the first benchmark that integrates both anomaly grounding and anomaly understanding. VAGU comprises over 7,567 real-world videos spanning 21 major anomaly categories (covering human criminal activities, natural disasters, animal-related injuries, traffic accidents, etc.), with an average duration exceeding 2,716 frames. In addition, we provide over 20,000 anomaly-related QA pairs to facilitate comprehensive anomaly understanding.

Furthermore, we innovatively propose the GtS (Glance then Scrutinize) framework based on VAGU, which achieves "coarse-grained temporal grounding → fine-grained anomaly comprehension → fine-grained anomaly grounding" through dynamic and static textual guidance under training-free conditions. This framework constructs anomaly grounding and anomaly understanding capabilities by leveraging existing multi-modal large language models(MLLMs).

Fig.~\ref{fig:performance compare} illustrates performance and inference speed comparisons between our framework and existing methods, while Fig,~\ref{fig:differ} delineates distinctions between our framework and other VAD paradigms. The proposed framework achieves an optimal balance between model performance and computational efficiency. For a detailed explanation of acceptable FPS and scores, please refer to the appendix.

Additionally, we introduce the JeAUG metric, which jointly quantifies semantic accuracy and grounding precision while incorporating video duration as a weighting factor. This enables more equitable evaluation of VAD capabilities across diverse data scenarios compared to conventional evaluation systems(AUC, AP and others). Extensive experiments on the proposed benchmark demonstrate that VAGU effectively supports complex VAD task assessments. The GtS framework exhibits significant performance advantages on VAGU, and JeAUG provides more comprehensive and fairer evaluation than traditional metrics.

Overall, our contributions are summarized as follows:
\begin{itemize}
	\item We have developed VAGU, a novel VAD benchmark that jointly addresses anomaly grounding and anomaly understanding. To the best of our knowledge, VAGU is the first large-scale benchmark that combines anomaly grounding and anomaly understanding, and also the first to provide an objective multiple-choice benchmark related to anomalies. Compared to existing datasets, our dataset is more comprehensive, challenging, and features higher annotation quality.
	\item We propose GtS, a training-free VAD framework that leverages text guidance to build capabilities for anomaly grounding and anomaly understanding on existing multi-modal large language models.
	\item Based on VAGU, we introduce an evaluation metric that jointly quantifies semantic accuracy and grounding precision.
	\item Extensive experiments have been conducted on the proposed VAGU, demonstrating the superiority of our benchmark, framework, and evaluation metric.
\end{itemize}

\section{Related Work}
\subsection{Traditional DNN-based VAD Paradigms}
Semi-supervised and weakly-supervised methods remain dominant in video anomaly detection (VAD). Semi-supervised approaches use self-supervised tasks to learn normal patterns~\cite{learn-28-lu2020few,learn-5-dong2020dual,learn-58-2017,synsyn-astrid2021synthetic,fu-35-wu2010chaotic}, such as auto-encoder reconstruction error and temporal prediction models~\cite{wang2019gods,wu2019deep}. Some recent works further improve robustness via multi-task learning~\cite{ECCV_2022_Jigsaw_Puzzles,paper3}, but still struggle with adapting to new scenarios, as minor viewpoint changes can reduce performance.

Weakly-supervised methods use video-level annotations, often with multiple instance learning or semantic priors for anomaly inference~\cite{weakly_Dual_2023,weakly_Generative_2022,vadclip_tsa,vadclip_vadclip}. While these improve detection accuracy, their reliance on manual annotation limits scalability in real-world deployments.

Unsupervised and open-set VAD methods~\cite{open-set-UBnormal,open-set-ding2022catching,open-set-zhu2022towards,open-set-zhu2023anomaly} seek to reduce annotation needs, but often perform poorly across varying scenarios due to architectural limitations. Across all supervised paradigms, a key challenge remains: limited anomaly understanding, which restricts deployment in complex real-world settings.

\begin{figure*}[h]
	\centering
	\includegraphics[width=\linewidth]{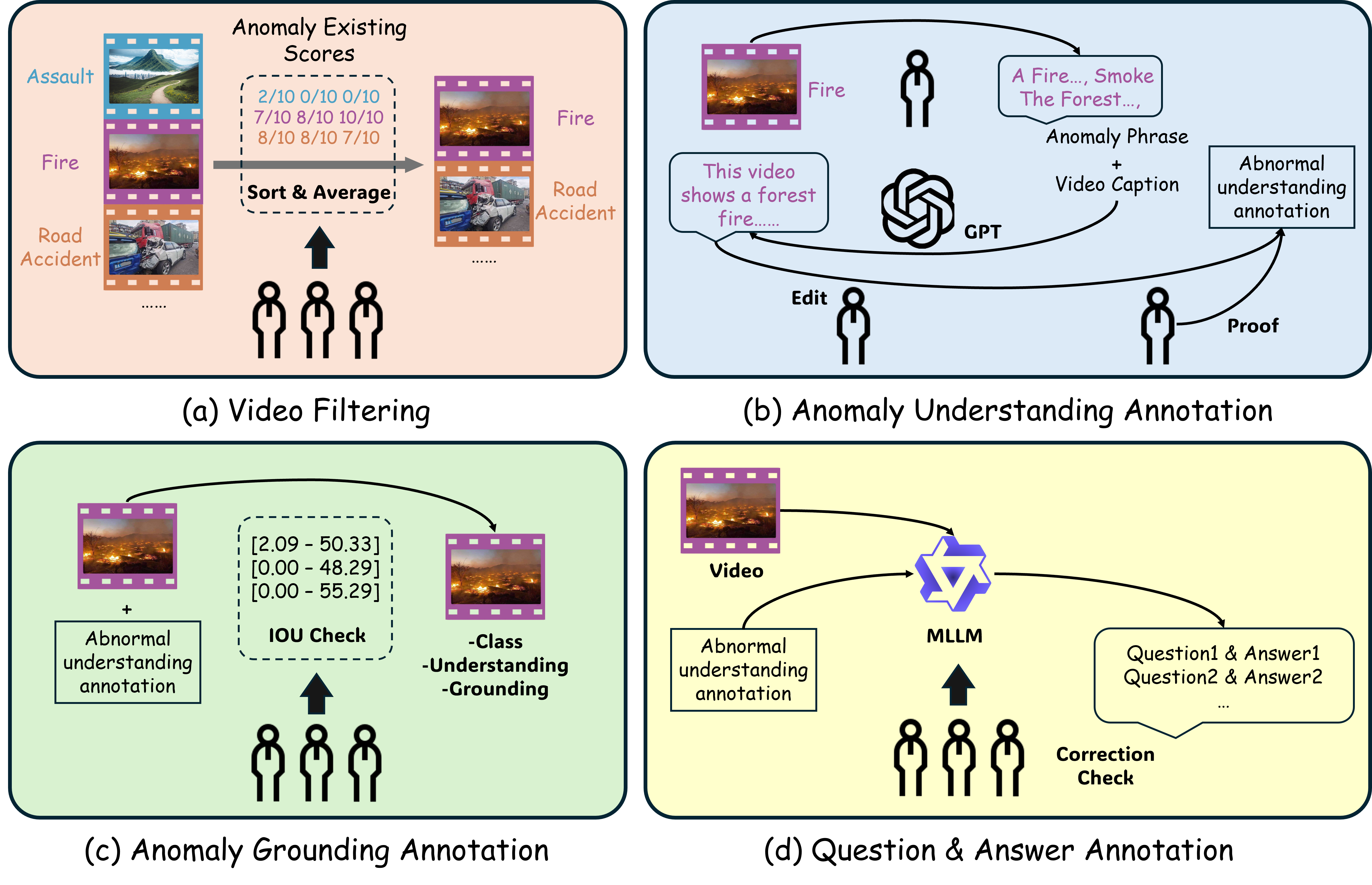}
	\caption{The annotation workflow of the VAGU dataset. Each sample is processed by at least three annotators collaboratively at every stage.}
	\label{fig:annotation}
\end{figure*}

\subsection{LLM-Based VAD Paradigms}
Recently, LLM-based VAD methods have advanced significantly. Vision-language models (VLMs) combine LLM reasoning with visual feature extraction, showing strong potential for VAD. Current approaches fall into two main categories:

The first uses external LLMs with frozen VLMs. Videos are segmented, described by VLMs, and processed by LLMs for semantic integration and detection~\cite{CUVA,HAWK,VERA,VAD-LLaMA,VANE-Bench,LAVAD,SUVAD,AnyAnomaly,FollowTheRules}. LAVAD~\cite{LAVAD} provides a training-free pipeline, while SUVAD~\cite{SUVAD} and FtR~\cite{FollowTheRules} introduce rule-mining and hallucination mitigation. Although these methods offer precise grounding and anomaly understanding, their frame/segment-wise processing incurs high computational costs, limiting real-time applications.

The second direction enhances VLMs via instruction tuning for interpretable predictions. Holmes-VAD~\cite{Holmes-VAD} uses QA datasets and temporal sampling. VAD-LLaMA~\cite{VAD-LLaMA} improves contextual modeling with LTC modules and a three-stage training strategy. CUVA~\cite{CUVA} adds a MIST selector for feature extraction, and VERA~\cite{VERA} generates instructional questions under weak supervision. While these methods boost detection performance, they require large domain-specific datasets for fine-tuning, leading to high computational cost and limited generalization.

\begin{table*}
	\centering
	\caption{A detailed comparison between the proposed VAGU dataset and other datasets. A.U. and A.G. mean Anomaly Understanding and Anomaly Grounding. The A.G. annotations in the CUVA and HIVAU-70k dataset are accomplished using VLM, which results in significant errors in application.}
	
	\begin{tabular}{cc|cc|cccc}
		\toprule
		
		\multirow{2}{*}{Dataset} & \multirow{2}{*}{Domain} & \multicolumn{2}{|c|}{Dataset Statistical Information}  & \multicolumn{4}{|c}{Dataset Annotation} \\
		
		\cline{3-8}
		
		& & Video Samples & Anomaly Categories & Audio & A.U. & A.G. & QA\\
		
		\midrule
		UCF-Crimes &Crime &1900 &13 & & &Frame/Human &\\
		XD-Violence &Violence &800 &6 &\checkmark & &Frame/Human &\\
		ShanghaiTech &Streetscape &437 &13 & & &BBbx/Human &\\
		UCSD Ped1 &Streetscape &70 &5 & & &BBbx/Human &\\
		UCSD Ped2 &Streetscape &28 &5 & & &BBbx/Human &\\
		CUHK Avenue &Streetscape &37 &5 & & &BBbx/Human &\\
		Street Scene &Traffic &81 &17 & & &BBbx/Human &\\
		CUVA &Multiple &1000 &11 &\checkmark &Caption/Human &Period/VLM &\\
		VANE-Bench &Multiple &325 &19 & & & &\checkmark\\
		HIVAU-70k &Multiple &5443 &15 & &Caption/Human &Period/VLM &\\
		\cline{1-8}
		
		VAGU(Ours) &Multiple &7567 &21 &\checkmark &Caption/Human &Period/Human &\checkmark\\
		
		\bottomrule
	\end{tabular}
	\label{tab:benchmark}
\end{table*}

\section{Proposed VAGU Benchmark}

In this section, we will first introduce the anomaly grounding and anomaly understanding tasks required by the VAGU dataset. Subsequently, we will describe the data collection process of VAGU and its semi-automatic annotation workflow with human verification. Finally, we will present detailed statistical information of the dataset along with comparative analyses against existing datasets.

\subsection{Task Definition}

\textbf{Video Anomaly Understanding.} 
This task comprises two objectives: anomaly classification and anomaly understanding.
In the anomaly classification subtask, the model is expected to output the category of the anomalous event present in the video, selected from a predefined anomaly category database.
In the anomaly understanding subtask, the model must analyze the given video content to comprehensively describe the subject, process, causes, and consequences potentially involved in the anomalous event.

\textbf{Video Anomaly Grounding.} 
This task requires the model to detect precise temporal intervals of anomalous events based solely on video data without external semantic information. This task's core constraints include: 
(1) \textit{Prompt Ambiguity:} The use of any video-level labels or fine-grained semantic labels provided in the dataset, except for the predefined anomaly type list, is strictly prohibited.
(2) \textit{Information Interactivity:} The model permits the incorporation of anomaly describe derived from the upper-level video anomaly understanding task as grounding evidence.

\subsection{Data Collection}

Compared to normal videos, those containing anomalous events are exceptionally scarce. We integrated existing datasets (CUVA~\cite{CUVA}, UCF-Crime\cite{ucf-crime}, and XD-Violence~\cite{xd-violence}) and collected over 12,000 videos potentially containing anomalies from major platforms such as YouTube, BiliBili, and TikTok. 
Through rigorous analysis and filtering, we curated 7,567 high-quality anomalous videos spanning 21 distinct anomaly categories across domains including human criminal activities, natural disasters, traffic accidents, and animal-inflicted injuries.

\subsection{Manual Annotation}

The construction of our VAGU dataset consists of three main steps: video filtering, semi-automatic anomaly annotation, and anomaly grounding. The process took about 650 hours and involved 20 annotators, with at least three annotators working on each video per stage. We collected videos from public datasets and online platforms, removed inappropriate content, and used a three-level quality review to select 7,567 high-quality anomaly videos. For annotation, we adopted a human-AI collaboration: annotators labeled key phrases, generated descriptions with vision-language models, and expanded them with ChatGPT. All annotations underwent multi-person cross-checks for accuracy and consistency. For anomaly grounding, at least three annotators independently marked the time intervals of anomalies, achieving consensus through IoU-based aggregation; ambiguous cases were iteratively re-annotated. Finally, we used advanced multimodal models to create multiple-choice QA tasks for each video, with annotators verifying correctness. Fig.~\ref{fig:annotation} illustrates this process, and we will present a more detailed annotation workflow in the appendix.

\subsection{Dataset Detailed Information}
The VAGU dataset comprises 7,567 high-quality anomaly videos, each annotated with triple labels for anomaly classification, understanding, and grounding. This dataset covers four major domains—human criminal activities, natural disasters, traffic accidents, and animal-related injuries—encompassing 21 fine-grained categories. Table.~\ref{tab:benchmark} provides a horizontal comparison with existing datasets through multi-dimensional metrics. Notably, certain categories (e.g., "Fire," "Arson," "Burning") exhibit semantic similarities; detailed definition criteria for differentiation can be found in the appendix.

Additionally, due to the limitations of existing VLMs, we encourage the use of multiple VLMs for ensemble-based completion of VAD tasks on the VAGU dataset.

\begin{figure*}[h]
	\centering
	\includegraphics[width=\linewidth]{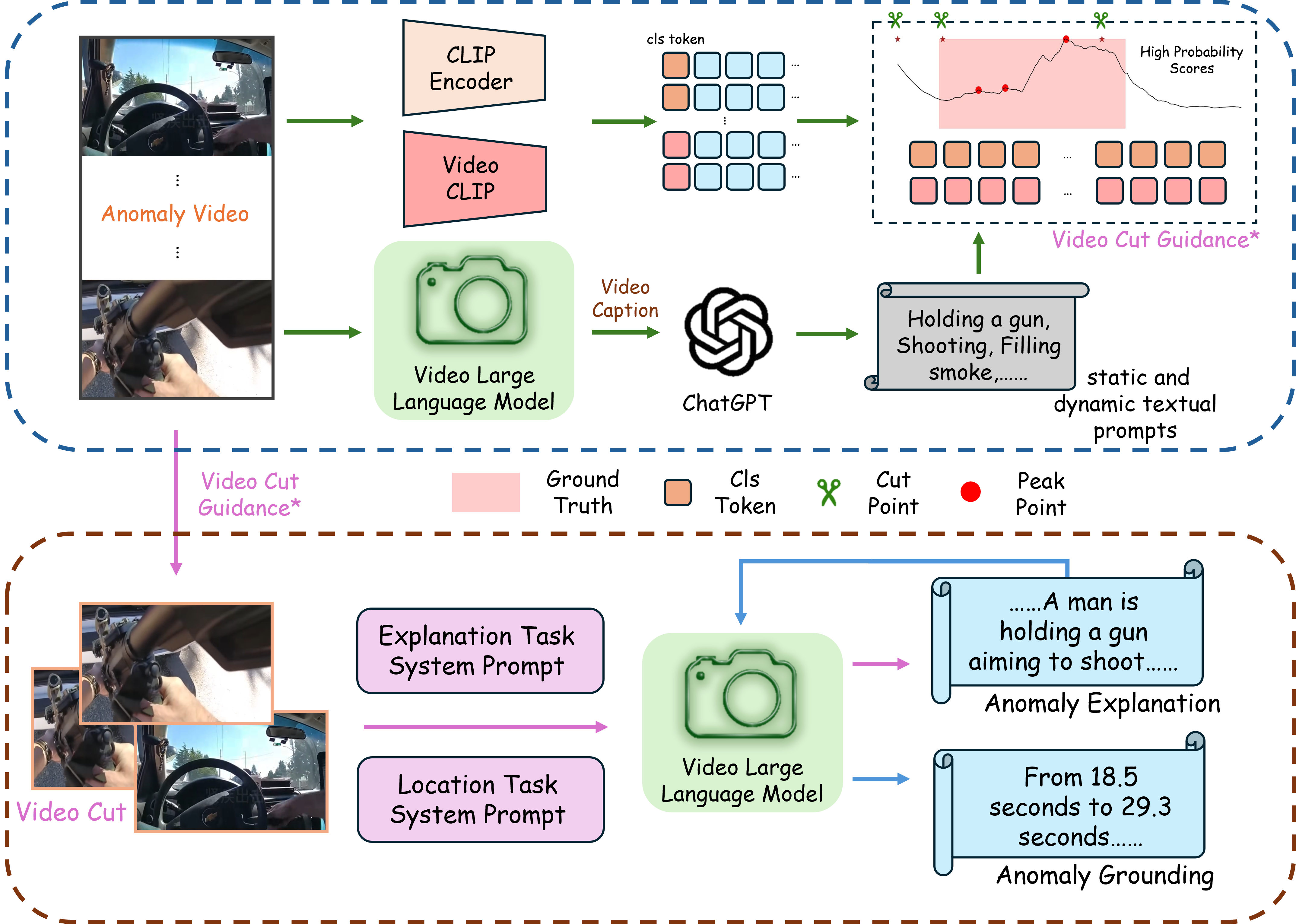}
	\caption{The flowchart of the GtS framework. In the first stage, GtS helps the model ground the time when the main event of the video occurs through dynamic and static text guidance. Then, in the second stage, GtS conducts fine-grained anomaly grounding and understanding based on the video segmentation in the previous stage.}
	\label{fig:model}
\end{figure*}

\section{Proposed Framework: Glance then Scrutinize}
In this section, we will introduce Glance then Scrutinize (GtS), an novel training-free VAD framework designed to address the three core challenges outlined previously. Built upon existing VLMs, our framework employs a dual textual guidance mechanism encompassing both dynamic and static contexts to ground potential anomalous video segments, thereby establishing comprehensive video anomaly grounding and understanding capabilities. Fig.~\ref{fig:model} illustrates the workflow of the proposed framework.

\subsection{Glance: Static \& Dynamic Text Guidance and Video Highlight Cut}
Our proposed framework accomplishes video anomaly grounding and understanding through a dual-phase pipeline. During the "Glance" phase, we initially leverage existing VLMs to generate video captions $~Cap_V~$ for input videos $~V_{input}~$. While current VLMs constrained by computational resources often misinterpret genuine anomalies as normal behaviors due to insufficient anomaly-specific token extraction, our experimental findings reveal that the generated descriptions maintain high accuracy in subject recognition. Building upon this insight, we construct multi-source textual inputs incorporating video captions, dataset-provided anomaly lists $~\mathcal{A}~$, and LLM-pregenerated contextual phrase banks $~\mathcal{B}_p~$ to drive GPT in parsing latent dynamic information (actions/events) and static information (subjects/scenes) from videos, thereby generating prompt lists for potential anomaly detection:
\begin{equation}
	\mathcal{PL}_s, \mathcal{PL}_d = {\rm LLM}(~Cap_V~, ~\mathcal{A}~, ~\mathcal{B}_p~).
\end{equation}

Crucially, it is noteworthy that the prompt lists returned by GPT may not inherently contain anomalous content, as this phase fundamentally aims to identify principal video subjects while filtering irrelevant background segments through semantic guidance.

We subsequently employ dual encoders (CLIP $~\Phi_{image}~$ and Video-CLIP $~\Phi_{video}~$) to perform feature encoding on video frames/segments, generating a temporal anomaly probability curve through cross-modal similarity computation:
%

\begin{equation}
	S_{s/d}(t) = \frac{
		\exp\left( \frac{1}{N} \sum_{n=1}^N \langle \Phi_{x}(pl_x^n),\, \Phi_{x}(V_t) \rangle \right)
	}{
		\sum_{t'} \exp\left( \frac{1}{N} \sum_{n=1}^N \langle \Phi_{x}(pl_x^n),\, \Phi_{x}(V_{t'}) \rangle \right)
	}
\end{equation}
where $x \in {{image}, {video}}$, and $N$ denotes the number of text descriptions in the corresponding branch; $pl_x^n$ represents the static or dynamic text descriptions; $V_t$ refers to the input frame or segment; $\Phi_x$ is the corresponding encoder; and $\langle \cdot, \cdot \rangle$ indicates the cosine similarity. $N$ is the number of static or dynamic text descriptions

Considering that events are continuous, we derive the overall similarity curve by combining the two similarity curves and apply the Savitzky-Golay filter to smooth the scores:
\begin{equation}
	S(t) = \frac{1}{Q}\sum_{p=-h}^{+h}(\alpha \cdot S_{s}(t) + (1-\alpha) \cdot S_{d}(t)) \cdot q_p,
\end{equation}
where ${q_p}/{Q}$ is the smoothing coefficient, determined through polynomial fitting using the least squares method, $\alpha$ is constant term.

Based on this curve, we implement a three-stage segmentation strategy: firstly detecting local extrema points within the similarity curve, secondly screening top-K candidate peaks according to inter-peak distances and magnitude thresholds:

\begin{equation}
	\mathcal{P} = \big\{t \in \left [0,T\right ) ~~\big|~~ S(t)=argmaxima(S)\big\},
\end{equation}
\begin{equation}
	\mathcal{P}^* = {\rm TopK}\Big(\big\{p \in \mathcal{P} ~~\big|~~ |p_i - p_j| > \theta, S(p_i)>= \tau\big\}\Big),
\end{equation}
where $\theta$ and $\tau$ are thresholds.

Finally, we perform dynamic window partitioning around the selected peaks while considering the total video duration, thereby segmenting the original video into high/low anomaly probability segments:
\begin{equation}
	\mathcal{H} = \bigcup_{p \in \mathcal{P}^*} [{\rm max}(0,p-\beta T), {\rm min}(T, p+\beta T)],
\end{equation}
where $\beta \in (0,1)$ controlling the window size proportion relative to $T$.

\subsection{Scrutinize: Fine-grained Anomaly Grounding and Understanding}

In the "scrutinize" phase, we establish a closed-loop integration of anomaly understanding and grounding through existing VQA and VTG models. For high-probability anomalous segments, we deploy the VQA model to detect and describe anomalous events based on predefined anomaly catalogs from dataset. 

To more precisely capture anomalous cues, we perform non-uniform sampling on each segmented video clip based on the similarity curve. Specifically, frames are selected where the normalized cumulative sum of $S(t)$ reaches equal intervals:

\begin{equation}
	\int_{a}^{t_i} S(t),dt = \frac{i}{N} \int_{a}^{b} S(t),dt, \quad i = 1, 2, \ldots, N
\end{equation}
where $S(t)$ is the similarity score curve, $\int_{a}^{b}$ denotes the segment interval, ${t_i}$ is the timestamp of the i-th sampled frame, and $N$ is the total number of sampled frames.

Furthermore, when performing anomaly detection on each non-initial video, we provide the model with the segmentation results from the previous video. This helps to maintain subject consistency in the subsequent integration process.

Compared to holistic video, the segmented analysis significantly enhances VLMs' capability to capture anomaly-relevant tokens, thereby effectively identifying subtle anomalies overlooked in full-length video processing. For low-probability segments, we employ the VQA model to generate captions while extracting latent anomaly-associated clues. 
Subsequently, we utilize a LLM to integrate the aforementioned captions, eliminating repetitive descriptions and those irrelevant to the anomalies, while establishing semantic connections among the segments. 
This process enables our framework to detect causally dependent anomalous behaviors (e.g., theft requiring sequential concealment and escape actions, arson involving combustible material placement and ignition procedures) through multi-segment evidence fusion.

Following comprehensive anomaly characterization, the GtS framework leverages VTG models for temporal grounding. 
By incorporating fine-grained semantic understanding as contextual prompts, our framework achieves superior grounding precision. 
This operational pipeline creates a mutually reinforcing and synergistic relationship between anomaly understanding and anomaly grounding tasks, fundamentally enhancing overall system performance through cognitive-visual alignment.

\section{Experiment}
\subsection{The Proposed JeAUG Metric}

The existing VAD evaluation metrics generally have the limitation of single-dimensional assessment. Some studies regard VAD as a VQA task and adopt text similarity measures such as ROUGE~\cite{ROUGE}, BLEU~\cite{BLEU}, and METEOR\cite{METEOR}, or generation quality evaluation metrics based on GPT series models, focusing on the assessment of anomaly semantic understanding. Another category of methods focuses on anomaly spatiotemporal grounding and mainly relies on traditional metrics in the field of computer vision such as AUC and AP. We propose an evaluation metric that jointly assesses anomaly understanding and detection (Joint Evaluation of Anomaly Understanding and Detection, JeAUG). 

This metric includes a dual-module evaluation framework: in the dimension of anomaly understanding, it guides external large language models (LLMs) to conduct multi-dimensional scoring on the semantic integrity and logical consistency of anomaly descriptions by constructing structured natural language prompt templates.
Specifically, we set up prompts to ask the LLM to score the results from four aspects: subject, scene, course of events, and impact. A score of 1/10 represents the lowest score, indicating that the response is almost entirely unrelated to the ground truth, while a score of 10/10 represents the highest score, indicating that the response is very appropriate in every aspect compared to the ground truth.
For the problem of subjective video boundary annotations in the dimension of anomaly grounding, we design an evaluation function that integrates video length weights based on human cognitive levels of abnormal events:

\begin{equation}
	F(IoU) = \frac{0.63}{ln10}ln(0.7 \cdot {\rm min}(\left\lfloor 10 \cdot IoU \right\rfloor, 7)+1)+0.5.
\end{equation}

Finally, we obtained the overall calculation equation for JeAUG:
\begin{equation}
	JeAUG = {\rm min}(\gamma \cdot {\rm F}(IoU), 1) \cdot Score_{A.U.} ,
\end{equation}
where $\gamma$ is the video length factor. For further rational discussion on JeAUG, please refer to the appendix.

\subsection{Implementation Details}
Our GtS framework employs different VLMs as the anomaly understanding model and anomaly grounding model. Meanwhile, we use CLIP-L/14 to encode video frames. All experiments were conducted using 14 A6000 GPUs, which took approximately 210 hours in total. For the LLM responsible for integration, we use Llama-3.1-8B. 
In Equation 4, $\alpha=0.4$, and $q_p/Q$ is obtained through least squares polynomial fitting, in Equation 6, $\gamma$ is the mean of all peak points, and $\theta$=total frame num / 12, in Equation 7, $\beta$=total frame num / 20. The above hyperparameters were derived by randomly sampling 100 videos from the anomalous videos that we discarded (not included in the evaluation videos) and calculating the proportion of anomalous timestamps.

\subsection{Quantitative evaluation of GtS}

\begin{figure*}[h]
	\centering
	\includegraphics[width=\linewidth]{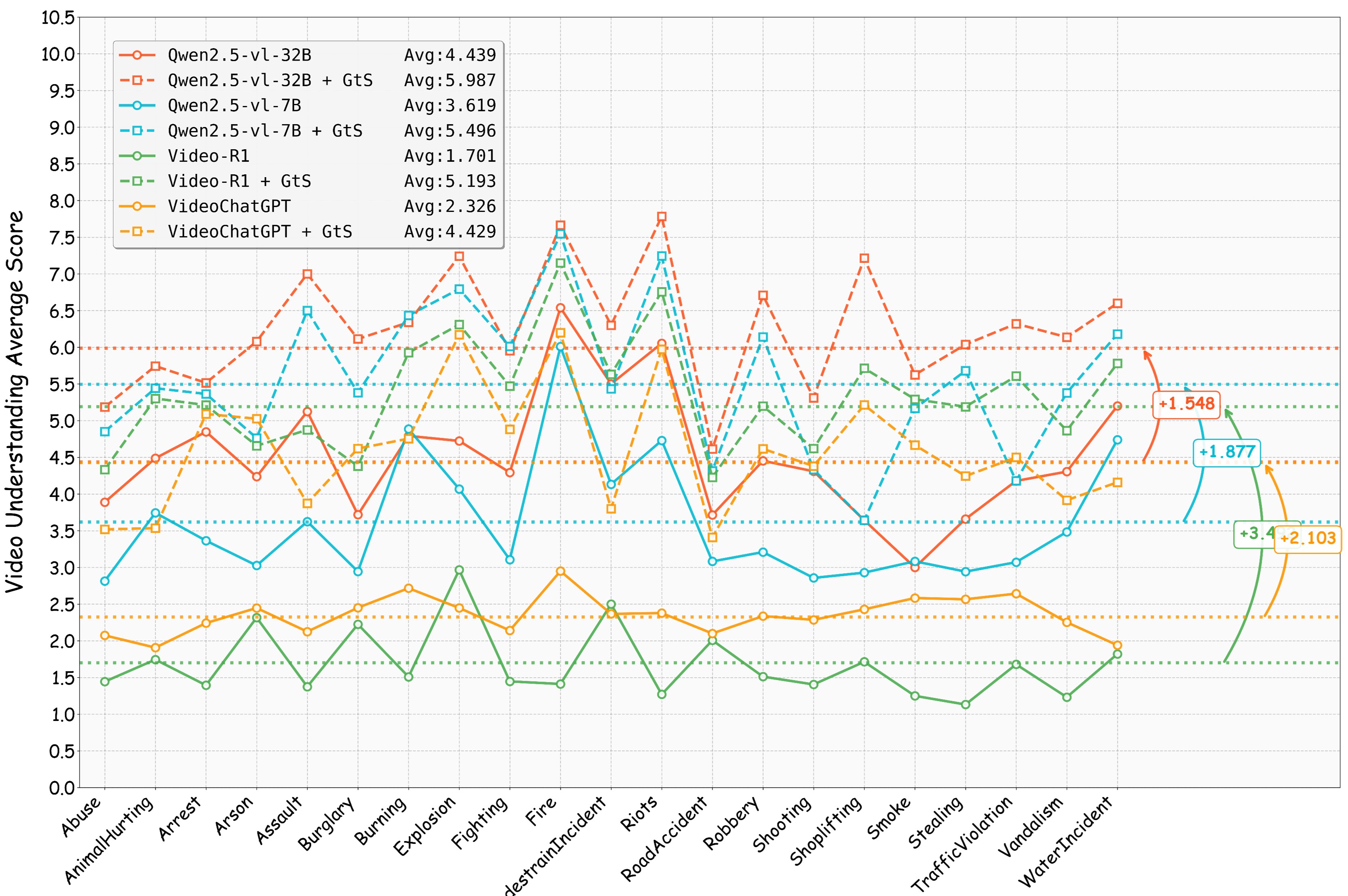}
	\caption{Fine-grained category comparison experiments on the video anomaly understanding sub-task.}
	\label{fig:class}
\end{figure*} 

We conducted a systematic comparative experiment on the proposed GtS framework on the VAGU dataset, and the results are shown in the Table.~\ref{table:result}. The experimental results indicate that GtS achieves a good balance between computational efficiency and detection accuracy. We note that due to the current limitations of VLMs, directly using VTG/VQA models to handle VAD tasks performs poorly. 

\begin{table}[h]
	\small
	\caption{The comparative experiments on the VAGU dataset. The GtS framework we proposed achieved the best balance between inference speed and detection accuracy. In the table, TC stands for TimeChat, and VT stands for VTimeLLM.}
	\centering
	\begin{tabular}{lcccc}
		\toprule
		Methods & A.U & JeAUG & QA & FPS \\
		\midrule
		\multicolumn{5}{c}{Frame/Segment-wise} \\		
		\midrule
		LAVAD & 5.52 & 4.47 & / & \textbf{0.24} \\
		SUVAD & \textbf{5.73} & \textbf{4.58} & / & 0.19 \\
		
		\midrule
		\multicolumn{5}{c}{Direct use of VQA/VTG models} \\
		\midrule
		VideoChatGPT + TC & 2.32 & 1.47 & 60.3\% & 229 \\
		VideoChatGPT + VT & 2.10 & 1.34 & $\uparrow$ & \textbf{286} \\
		Video-XL + TC & 2.31 & 1.55 & 58.6\% & 192 \\
		Video-XL + VT & 2.13 & 1.38 & $\uparrow$ & 201 \\
		Qwen2.5-VL-7B + TC & 3.61 & 2.28 & 68.0\% & 185 \\
		Qwen2.5-VL-32B + TC & \textbf{4.43} & \textbf{2.78} & 71.9\% & 95 \\
		Video-R1 + TC & 1.70& 1.08 & \textbf{80.0\%} & 112 \\
		
		\midrule
		\multicolumn{5}{c}{Ours GtS} \\
		\midrule
		Qwen2.5-VL-7B + TC* & 5.50 & 4.04 & 73.5\% & 61 \\
		Qwen2.5-VL-32B + TC* & \textbf{5.99} & \textbf{4.30} & 76.8\% & 36 \\
		Video-R1 + TC* & 5.19 & 3.69 & \textbf{88.9\%} & 42 \\
		VideoChatGPT + TC* & 4.42 & 3.26 & 65.1\% & \textbf{71} \\
		
		\bottomrule
	\end{tabular}
	\label{table:result}
\end{table}

To further verify the effectiveness of the framework, we conducted a fine-grained category analysis on the anomaly understanding sub-task. As shown in Fig.~\ref{fig:class}, compared with the baseline, GtS shows a significant improvement in anomaly categories that require more anomaly clues (such as Arrest, Arson, Riots, Shoplifting, etc.), which fully demonstrates the superiority of GtS. 

\subsection{Ablation \& Case Study}

\begin{figure}[h]
	\centering
	\includegraphics[width=\linewidth]{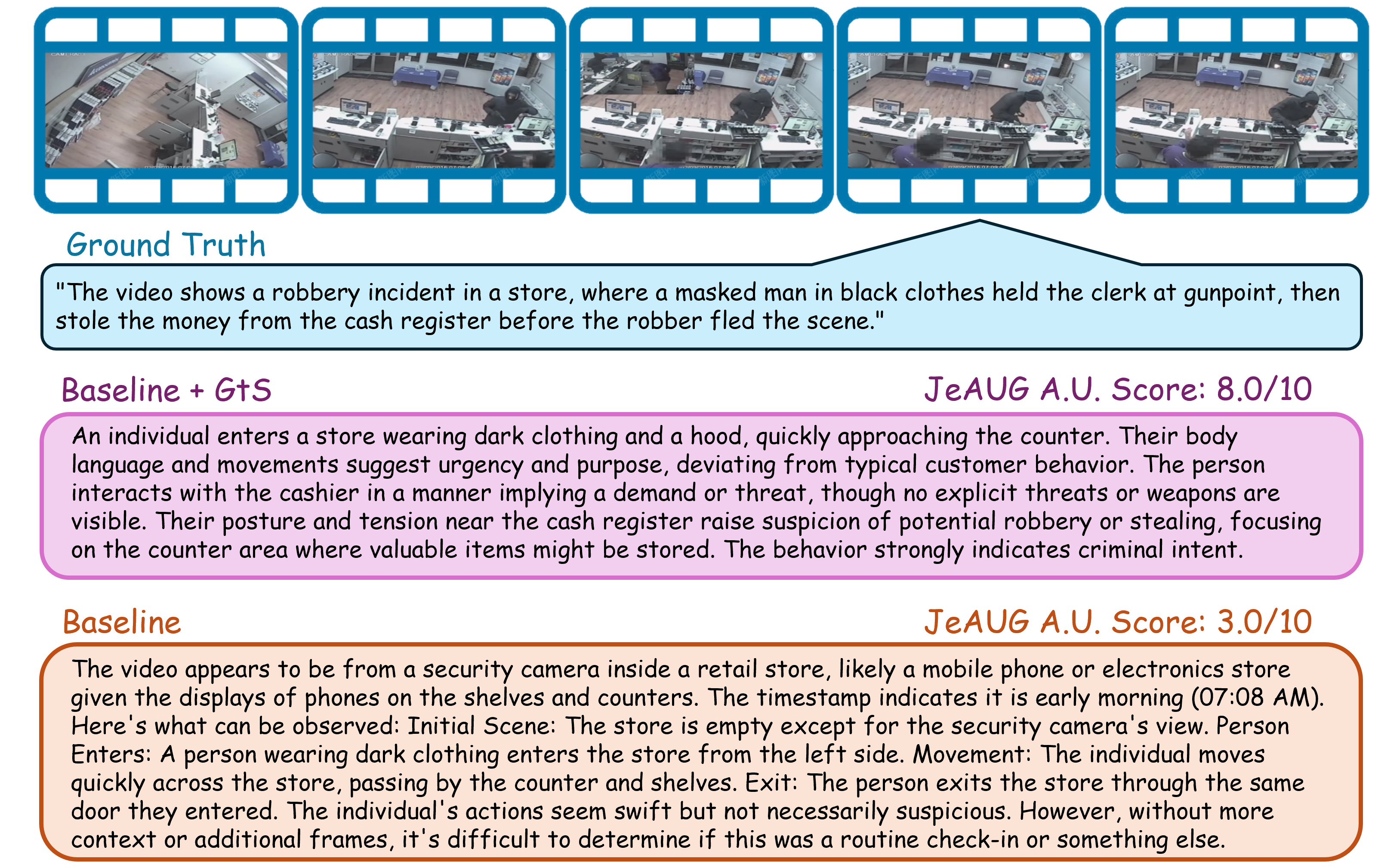}
	\caption{The case study of the GtS framework on the VAGU dataset.}
	\label{fig:case}
\end{figure}

In Fig.~\ref{fig:case}, we present a case study of the GtS framework on the VAGU dataset. It can be clearly observed that GtS accurately identified the primary anomaly event and provided a detailed analysis of the entire process. Although the output still shows some hallucination effects and contains redundant descriptions, the training-free VAD framework of GtS undoubtedly exhibits tremendous potential.

\begin{table}[h]
	\small
	\centering
	\caption{The experiment on the impact of dynamic and static text.}
	\begin{tabular}{lc}
		\toprule
		Model &JeAUG A.U. Scores\\
		\toprule
		Qwen2.5-VL-7B + GtS &\textbf{5.50}\\
		- dynamic text guidance & 5.27\\
		- static text guidance & 5.30\\
		- integral non-uniform sampling & 5.38\\
		- using contextual understanding & 5.41\\
		\toprule
	\end{tabular}
	\label{table:text}
\end{table}

In addition, we also investigated the impact of modules such as dynamic texts and integral non-uniform sampling on the anomaly detection performance of the GtS framework, and the results are shown in Table~\ref{table:text}. It can be seen that each module enhances the anomaly detection capability of the GtS framework.

\subsection{Additional Experiments and Analyses}
Due to space limitations, we present additional experimental results in the appendix, including the human preference rationality experiments for JeAUG and a detailed description of the localization branch process. Furthermore, we provide a comprehensive inference procedure with prompts. In addition, we include further details about our dataset, category distinctions, and ethical guidelines.

\section{Conclusion}
In this paper, we introduce VAGU, the first benchmark in the VAD field that simultaneously considers video anomaly understanding and grounding. Compared with existing datasets, VAGU is more comprehensive, more challenging, and has higher annotation quality, with all annotations undergoing multiple rounds of manual checks. We believe that VAGU can significantly encourage the exploration and development of VAD methods, especially LLM-based VAD methods, and lay the foundation for the development of VAD tasks in the era of LLMs. Additionally, we propose a VAD framework GtS, which uses dynamic and static text guidance. GtS can follow the general setting of VAD, first grounding the approximate time when the subject event occurs in the video using only the abnormal list provided by the dataset, and then conducting fine-grained anomaly understanding and grounding, achieving a great balance between computational cost and detection performance. We also propose a joint metric, JeAUG, for evaluating the performance of anomaly understanding and grounding, which can more comprehensively and fairly assess model performance. Experimental results show that the introduction of VAGU brings new research directions to VAD.

\bibliographystyle{IEEEtran}
\bibliography{paper5}

\appendices

\section{Rationale for JeAUG}
\subsection{JeAUG A.U. Score}

In the anomaly understanding task, we aim for models to generate video descriptions that are accurate, clear, concise, and coherent. While numerous evaluation metrics exist for semantic similarity assessment (e.g., ROUGE, BLEU, METEOR), the descriptions of infrequent anomalous events often vary widely in wording. These conventional metrics are prone to biases from sentence length, synonym use, and word order variations. We divided each category in the VAGU dataset into five parts (rounded down) to form five subsets. We evaluate the Coefficient of Variation (CV) for four metrics, the results are shown in the Table.~\ref{table:cv}. 

\begin{table}[ht!]
	\caption{Comparative experiments on the Coefficient of Variation for evaluation metrics.}
	\centering
	\label{table:cv}
	\begin{tabular}{ccccc}
		\toprule
		&ROUGE &BLEU &METOR &JeAUG A.U.\\
		\toprule
		CV &3.12 &31.72 &3.33 &2.32 \\
		\toprule
	\end{tabular}
\end{table}

The results demonstrate that our proposed metric exhibits greater stability when processing abnormality-related videos.

Given that the JeAUG takes into account both anomaly grounding and understanding evaluations, and the VAGU dataset supports the integrated application of multiple VLM models, to avoid assessment imbalances caused by excellent performance in one dimension but weak performance in the other (such as precise grounding but failed semantic parsing), we set the minimum effective score for the anomaly understanding dimension to 1 instead of 0.

\subsection{JeAUG A.G. Score}

\begin{figure}[h]
	\centering
	\includegraphics[width=\linewidth]{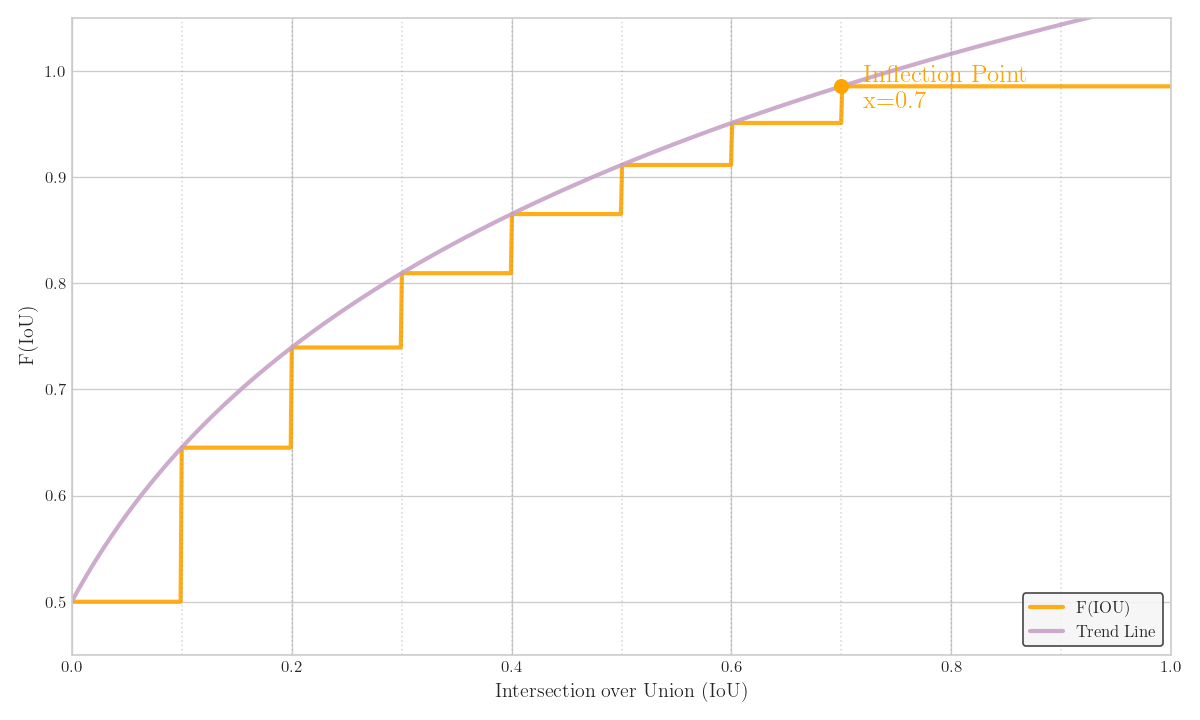}
	\caption{Visualization of the function $F(IoU)$ in JeAUG A.G.}
	\label{fig:jeaug}
\end{figure}

First, the grounding of anomalous events is highly subjective. In real-world scenarios, it is difficult to pinpoint the exact moment an abnormal event occurs. For example, in a car accident, the starting point could range from seconds to tens of seconds before the collision. To rationalize evaluation metrics for abnormal event grounding, we aligned them with human preferences. Specifically, we invited over a dozen evaluators to ground anomalies in diverse videos. We then calculated the pairwise Intersection over Union (IoU) ratios of their results for the same video and took the average across all samples as the human preference consensus. Experimental results showed a value around 0.7. Accounting for human grading preferences and the possibility of accurate abnormality comprehension but failed grounding, we designed the following piecewise function:

\begin{equation}
	F(IoU) = \frac{0.63}{ln10}ln(0.7 \cdot {\rm min}(\left\lfloor 10 \cdot IoU \right\rfloor, 7)+1)+0.5,
\end{equation}
the visualization of the function is provided in Fig.~\ref{fig:jeaug}.

\begin{figure}[h]
	\centering
	\includegraphics[width=\linewidth]{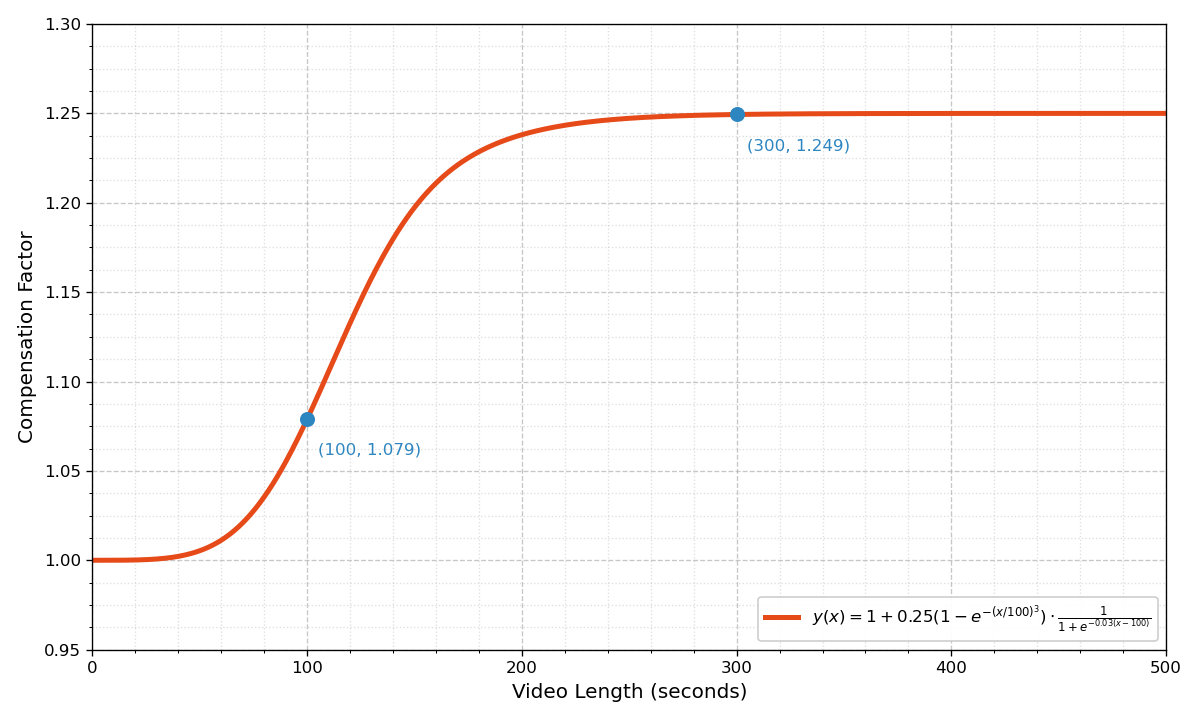}
	\caption{Visualization of the video length coefficient $\gamma$.}
	\label{fig:gamma}
\end{figure}

During experiments, we observed that grounding anomalies in longer videos is generally more challenging than in shorter videos. This is attributed to the fact that under the premise of general VTG models extracting a fixed number of frames, longer videos contain proportionally less abnormality-related information, thereby increasing grounding difficulty. Thus, we introduced a video length compensation factor $\gamma$, defined as follows:

\begin{equation}
	\gamma = G(T) = 1 + 0.25\left(1 - e^{-(T/100)^3}\right)\left(\frac{1}{1 + e^{-0.03(T-100)}}\right),
\end{equation}
the visualization of the function is provided in Fig.~\ref{fig:gamma}.

Finally, the overall JeAUG formula is defined as:
\begin{equation}
	JeAUG = {\rm min}(\gamma \cdot F(IoU), 1) \cdot Score_{A.U.} 
\end{equation}

Our study aims to achieve collaborative optimization of video anomaly understanding and video anomaly grounding performance, avoiding performance imbalances caused by single-task dominance. Under conditions lacking video-level category label supervision, we evaluated mainstream Vision-Language Models (VLMs) on the VTG task, with experimental results shown in Table.~\ref{table:vtg}. 

\begin{table}[ht!]
	\caption{Experimental results of mainstream VLM on the video anomaly grounding task using the VAGU dataset.}
	\label{table:vtg}
	\centering
	\begin{tabular}{cc}
		\toprule
		Method & IoU\\
		\toprule
		mPLUG-Owl & 12.6\%\\
		Video-LLaMA & 7.3\%\\
		VideoChatGPT & 6.1\%\\
		TimeChat & 14.8\%\\	
		\toprule
	\end{tabular}
\end{table}

Data analysis revealed that current VLMs perform poorly on anomaly video grounding task. Therefore, we designed the video length factor $\gamma$ and set 0.5 as the starting point for the ${\rm F}(IoU)$ function. This design effectively alleviates evaluation distortion caused by accurate semantic understanding but grounding deviations through parameterized task weight distribution.

\begin{figure*}[h]
	\centering
	\includegraphics[width=\linewidth]{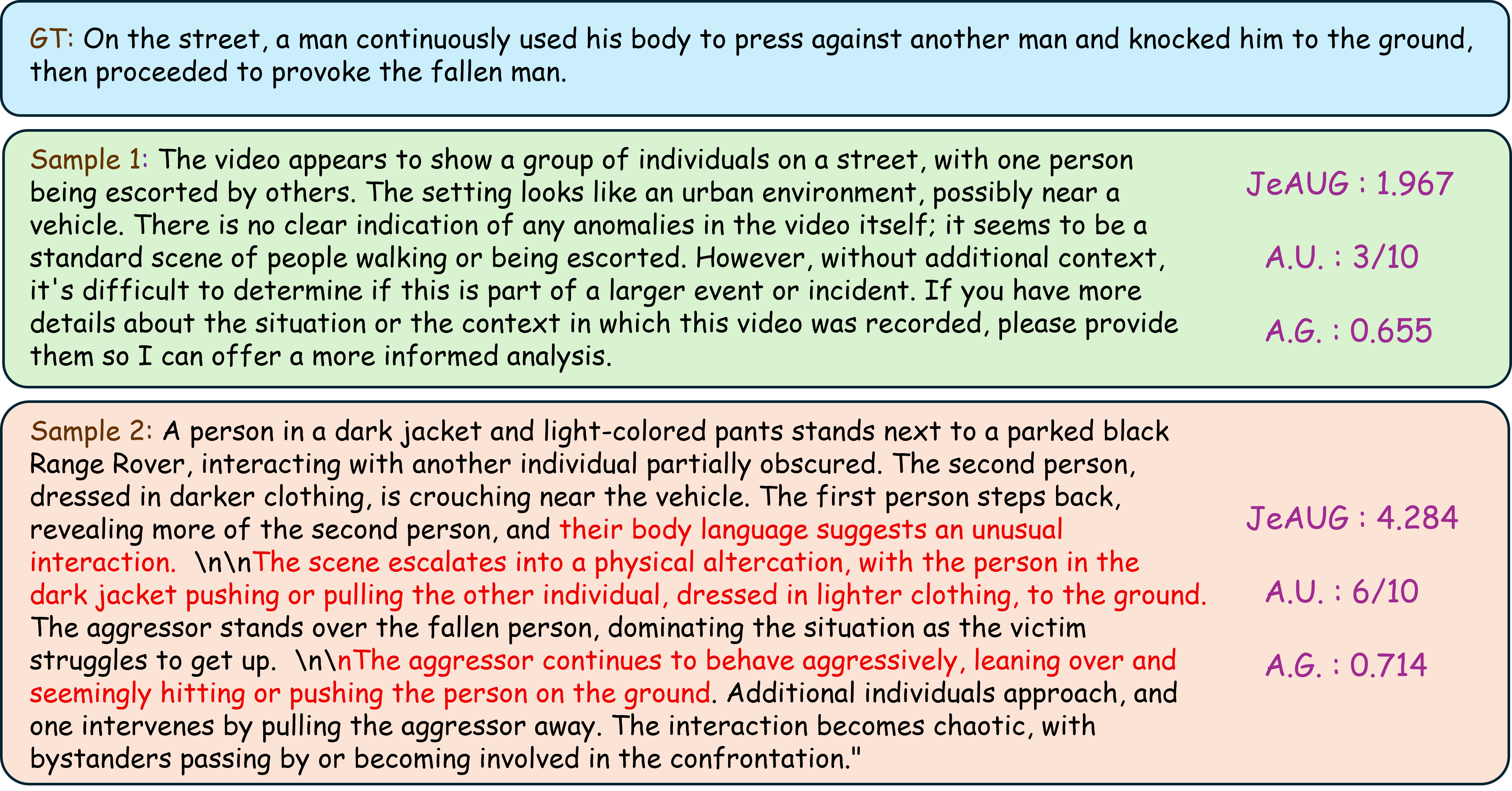}
	\caption{A schematic diagram of the scores assigned by different models on the same sample illustrates that a JeAUG score exceeding 3 indicates that the model can effectively understand the anomalous events occurring in the video.}
	\label{fig:sup1}
\end{figure*}

\section{Hyperparameter Selection For JeAUG}

In the design of our JeAUG evaluation metric, the integration of the understanding and grounding branches involves several manually defined hyperparameters, such as the video length compensation factor $\gamma$ and the weighting coefficients of the $F(IoU)$. It is important to emphasize that these hyperparameters were not tuned for any specific dataset; rather, they were set to ensure that the evaluation outcomes of JeAUG align as closely as possible with human judgment. To this end, we conducted a human preference alignment experiment.

Specifically, we invited 10 participants to independently perform anomaly understanding and anomaly grounding on 50 randomly selected videos containing anomalous events, following the same procedure as the MLLM. The experimental results show that, for the same anomalous event, the lowest pairwise IoU among the 10 participants is approximately 0.7. In addition, considering the evaluation results of general large models’ anomaly understanding capabilities discussed in the main text, as well as the overall suboptimal grounding performance reported in Table 2, we draw the following conclusions:
\begin{itemize}
	\item Human consensus on anomaly grounding (i.e., achieving full grounding score) corresponds to an IoU of approximately 0.7.
	\item Improvements in coarse grounding accuracy are more significant than those in fine grounding (i.e., the performance gain from increasing IoU from 0.5 to 0.6 is much greater than from 0.9 to 1.0).
	\item There exist cases where anomaly understanding is successful even when anomaly grounding completely fails.
\end{itemize}

Based on these findings, we designed the scoring function as follows:
\begin{itemize}
	\item When the IoU reaches 0.7, the grounding score is assigned the maximum value.
	\item An initial score is set to account for cases where anomaly grounding fails but anomaly understanding succeeds.
	\item A logarithmic-like incremental scheme is adopted, so that score increases gradually slow down as the IoU improves.
	\item The function is piecewise-defined according to human preferences.
\end{itemize}

On this basis, the relevant hyperparameters were determined, and the JeAUG metric was implemented accordingly.

\section{Distinguishable Thresholds for A.U. and A.G.}

As shown in \textbf{Fig.~1 of the main text}, we define the threshold criteria for acceptable anomaly understanding ability (A.U.) and anomaly grounding ability (A.G.) as JeAUG score $\geq$ 3 and FPS $\geq$ 30, respectively. For inference speed, we reference the frame rates of mainstream video formats and align with the standard human viewing speed, setting the acceptable FPS to 30.

Regarding video understanding ability, Fig.~\ref{fig:sup1} presents representative case analyses. It is evident that when the JeAUG score falls below 3, the model exhibits poor performance in both understanding and grounding anomalous events, often failing to provide accurate judgments. In contrast, when the JeAUG score exceeds 3, the model generally demonstrates satisfactory performance in both understanding and grounding anomalies, despite occasional misjudgments. Therefore, we set JeAUG score $\geq$ 3 as the lower bound of the acceptable range.

In summary, the distinguishable thresholds for AU and AG are defined as JeAUG score $\geq$ 3 and FPS $\geq$ 30, respectively, serving as the basic criteria for evaluating model performance.

\begin{figure*}[h]
	\centering
	\includegraphics[width=\linewidth]{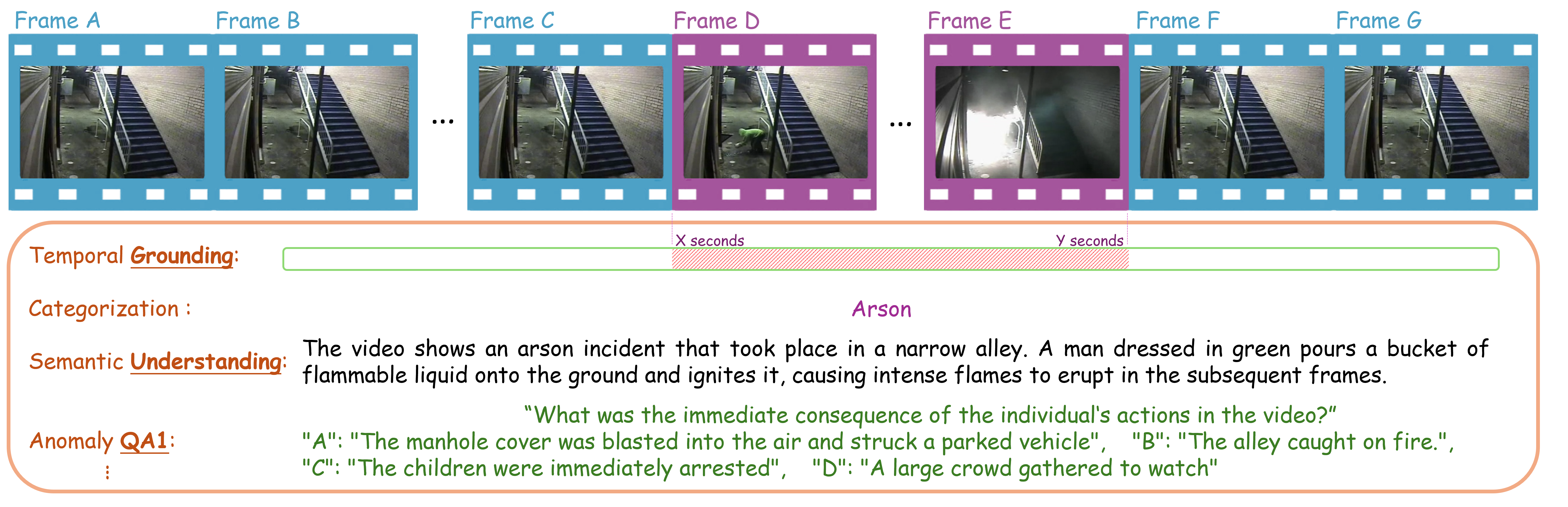}
	\caption{An annotation example in our proposed benchmark. Each sample includes four manually validated annotations: anomaly category, anomaly grounding, anomaly description, and anomaly QA. This dataset can be used for VAD, VQA, or VTG tasks.}
	\label{fig:data sample}
\end{figure*}

\section{Annotation Sample Description}

Fig.\ref{fig:data sample} presents a detailed annotation example from our proposed VAGU dataset. Specifically, each anomalous video is annotated with a set of four elements: anomaly category, anomaly event grounding, anomaly event understanding, and 2–5 multiple-choice questions (QA) related to the anomalous content. For normal videos, the category is labeled as “Normal”; these samples include content understanding annotations and 2–5 multiple-choice questions regarding the video content, but do not contain anomaly event grounding annotations.

Based on this design, the VAGU dataset supports a variety of mainstream Video Anomaly Detection (VAD) tasks:
\begin{itemize}
	\item \textbf{Categorization} and \textbf{Temporal Grounding}: Supports traditional weakly supervised VAD tasks.
	\item \textbf{Normal videos only}: Supports traditional semi-supervised VAD tasks.
	\item \textbf{Semantic Understanding}: Meets the current research demand for understanding anomalous events.
	\item Joint \textbf{Semantic Understanding }and \textbf{Temporal Grounding}: Supports our proposed joint VAD task.
	\item \textbf{Anomaly QA}: Provides an objective standard for evaluating large models’ anomaly understanding capability.
\end{itemize}

\begin{table*}[htbp]
	\centering
	\caption{Supported VAD Tasks for Different Annotation Combinations}
	\begin{tabular}{|l|c|c|c|c|c|}
		\hline
		\textbf{Annotation Combination}           & \textbf{WVAD} & \textbf{SVAD} & \textbf{A.U.} & \textbf{Joint A.U. and A.G.} & \textbf{Anomaly QA} \\
		\hline
		Anomaly Category + Temporal Grounding     & \checkmark& \checkmark&   &   &   \\
		Normal videos only & &\checkmark &  &   &   \\
		Semantic Understanding & \checkmark&   & \checkmark&   & \checkmark\\
		Semantic Understanding + Temporal Grounding & \checkmark& \checkmark& \checkmark& \checkmark& \\
		Anomaly QA &   &   &   &   & \checkmark\\
		\hline
	\end{tabular}
	\label{table:sup1}
\end{table*}

It is worth noting that the VAGU dataset contains both normal and anomalous videos, facilitating large-scale model fine-tuning and other training tasks, thereby greatly enhancing its practical utility. Table.~\ref{table:sup1} provides a detailed summary of the types of tasks supported by different annotation combinations within the dataset.

\section{Independent Evaluation of Video Anomaly Understanding, Video Anomaly Grounding, and Anomaly Question Answering}

\begin{figure*}[h]
	\centering
	\includegraphics[width=\linewidth]{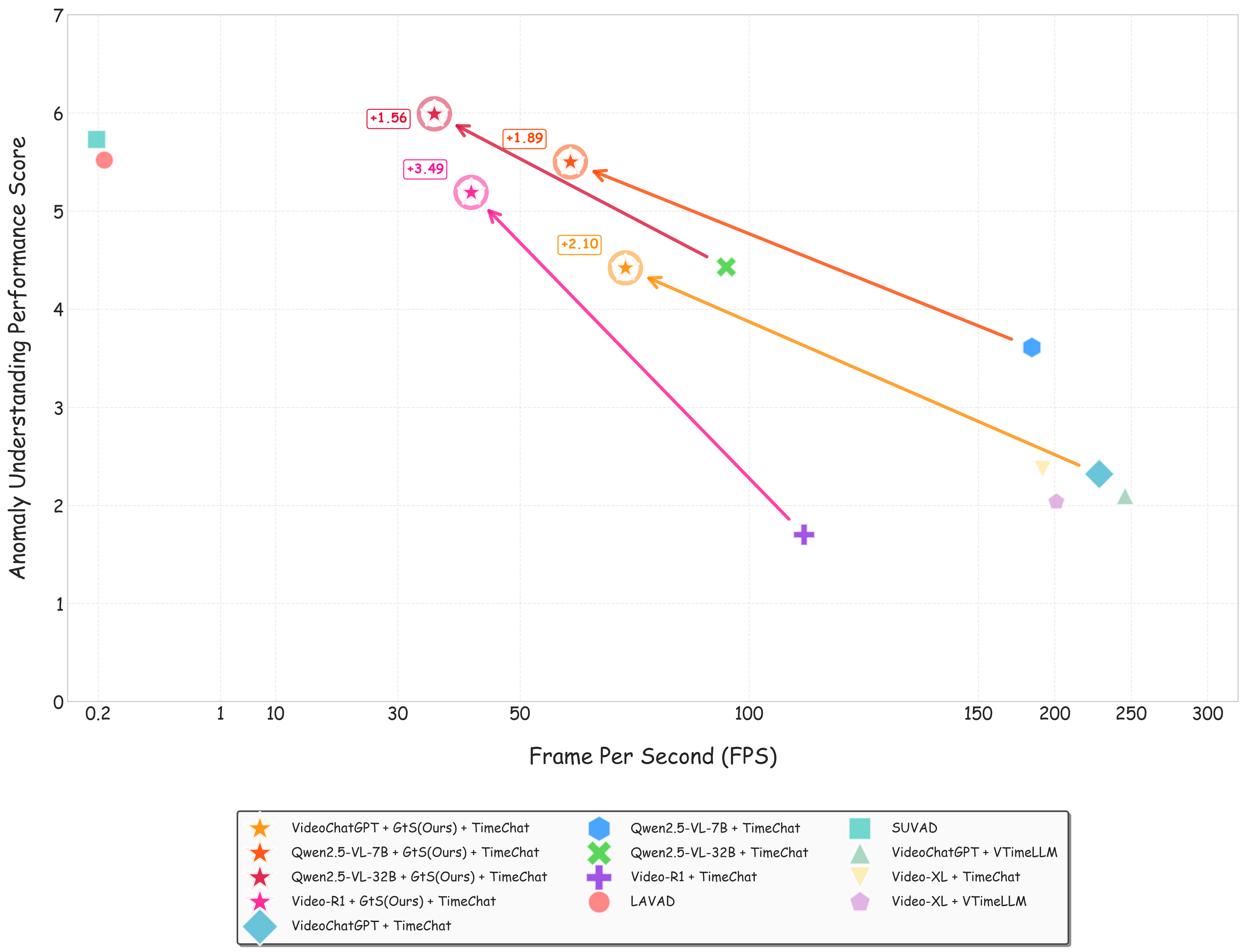}
	\caption{The comparison of Anomaly Understanding performance between various baselines and the models incorporating the GtS framework.}
	\label{fig:sup2}
\end{figure*}

To further assess and analyze the performance of our proposed GtS framework on each subtask, we conducted independent evaluations and visual analyses for video anomaly understanding, video anomaly grounding, and anomaly question answering (QA), respectively.

\begin{figure}[h]
	\centering
	\includegraphics[width=\linewidth]{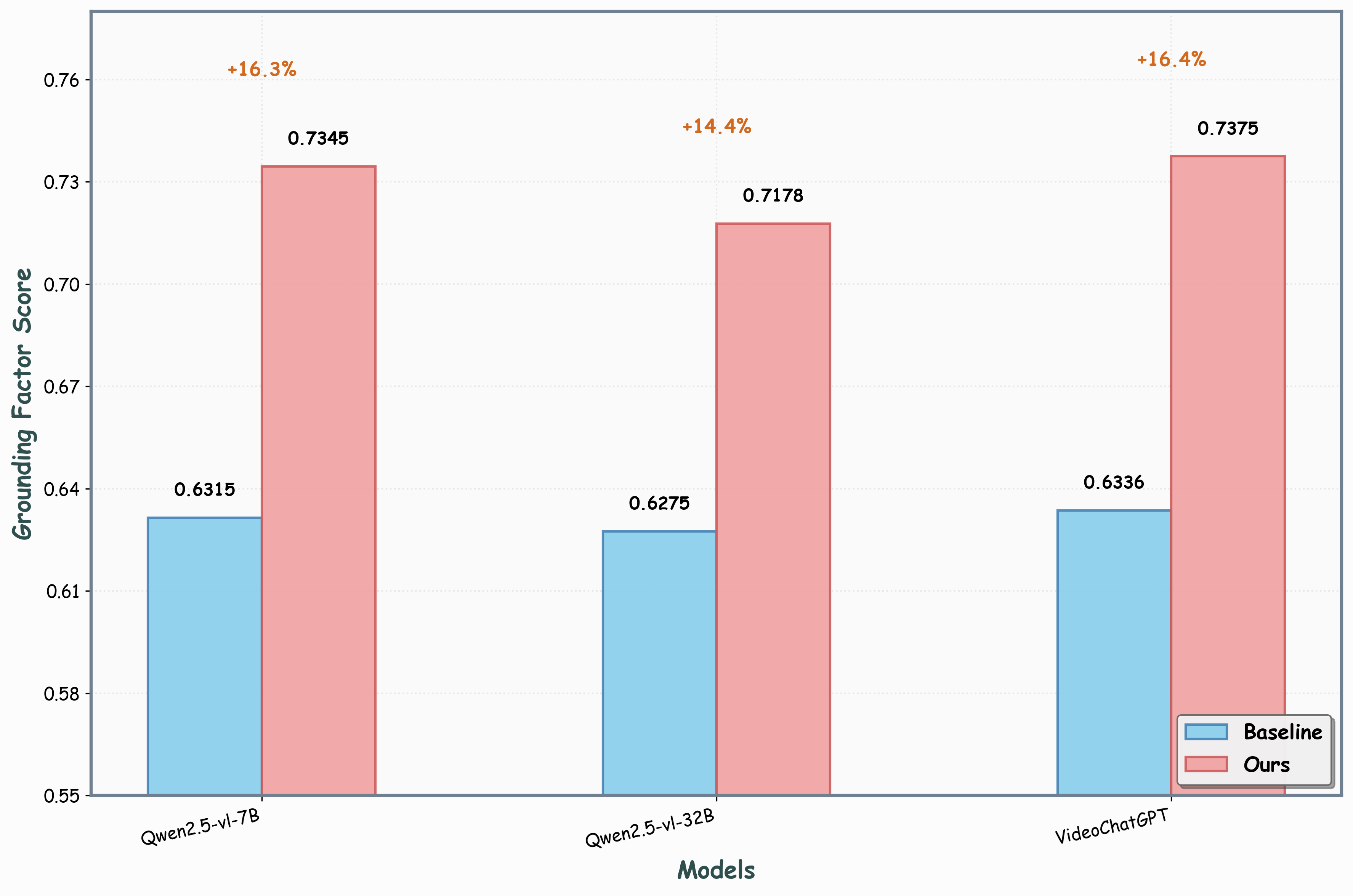}
	\caption{This bar chart illustrates the improvement in grounding factor performance achieved by our proposed method across three different vision-language models. The grounding factor is a key metric within our proposed JeAUG benchmark, used to evaluate the accuracy of anomaly event grounding by the models.}
	\label{fig:sup3}
\end{figure}

\begin{figure}[h]
	\centering
	\includegraphics[width=\linewidth]{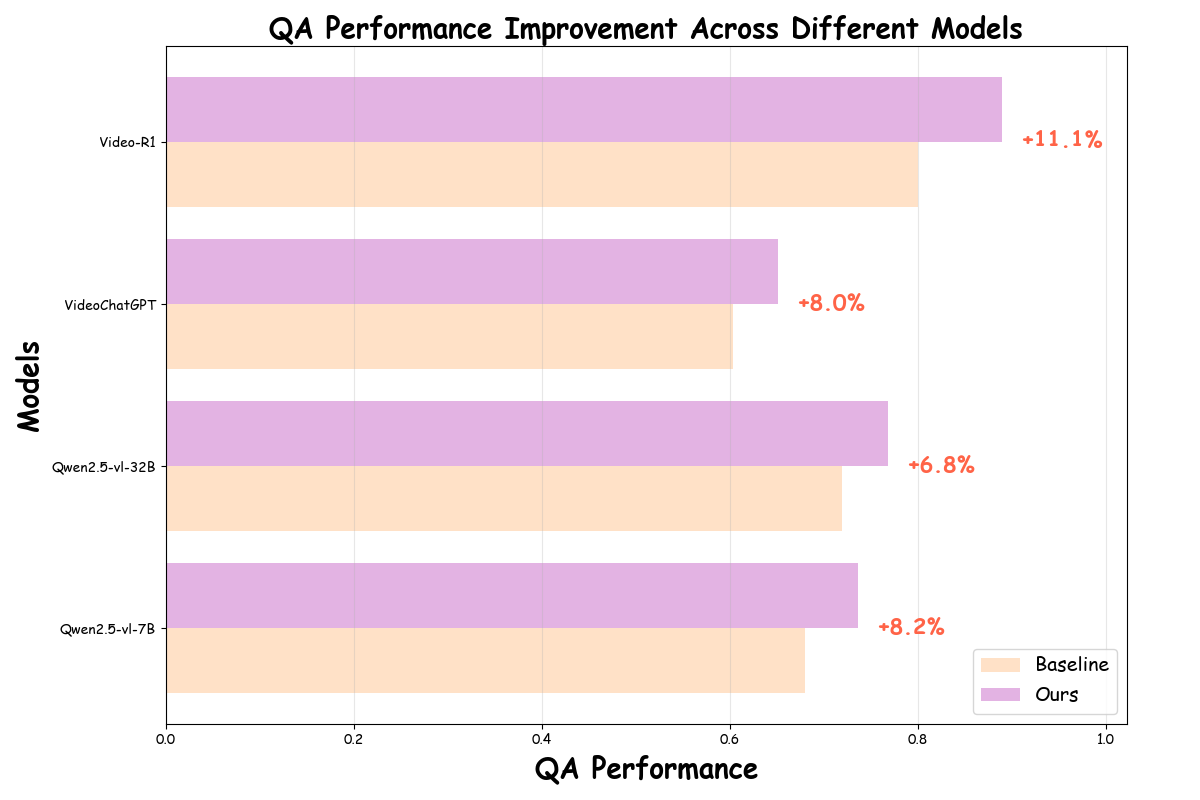}
	\caption{This figure shows the Anomaly QA performance improvement across different models, comparing the baseline and our GtS.}
	\label{fig:sup4}
\end{figure}

Fig.\ref{fig:sup2} presents a comparison of Anomaly Understanding performance between various baselines and the models incorporating the GtS framework. Fig.~\ref{fig:sup3} further illustrates the improvement in anomaly grounding capability brought by the GtS framework, with the grounding factor description $\gamma \cdot F(IoU)$ from the JeAUG benchmark used as the evaluation criterion. Fig.~\ref{fig:sup4} presents the performance of the GtS framework on the anomaly question answering (QA) task.

Analysis of the experimental results reveals that the GtS framework yields significant performance gains across all subtasks, with the most notable improvements observed in the video anomaly understanding task. Specifically, for models with relatively small parameter sizes (e.g., VideoChatGPT, Qwen-2.5-VL-7B), GtS enables more accurate detection of anomalous cues, thereby facilitating better comprehension of abnormal phenomena in videos. For larger models with strong video understanding capabilities (e.g., Qwen-2.5-VL-32B), the GtS framework further enhances the models' ability to provide detailed descriptions of anomalous events.

It is worth noting that models with reasoning capabilities, such as Video-R1, often experience performance degradation when directly reasoning over the entire video due to interference from irrelevant frames. In contrast, the GtS framework can effectively guide the model to focus on key content, thereby achieving objective performance improvements. In the abnormal event grounding task, GtS is able to provide explicit anomaly cues, which significantly enhances grounding accuracy. Although the performance gains of GtS on the anomaly QA task are less pronounced than those observed in the previous two tasks, the overall improvement still exceeds 10\%. We attribute this to the limitations of current VQA/VTG models in handling open-ended questions, whereas their performance is relatively better on closed-ended questions—a point that has been discussed in the Introduction section of the main text. Nevertheless, the GtS framework still brings substantial performance improvements. Moreover, considering that anomaly QA tasks have limited practical applications in real-world scenarios, the remarkable improvements achieved by the GtS framework in video anomaly understanding and anomaly grounding tasks further underscore its practical value.

\begin{figure*}[h]
	\centering
	\includegraphics[width=\linewidth]{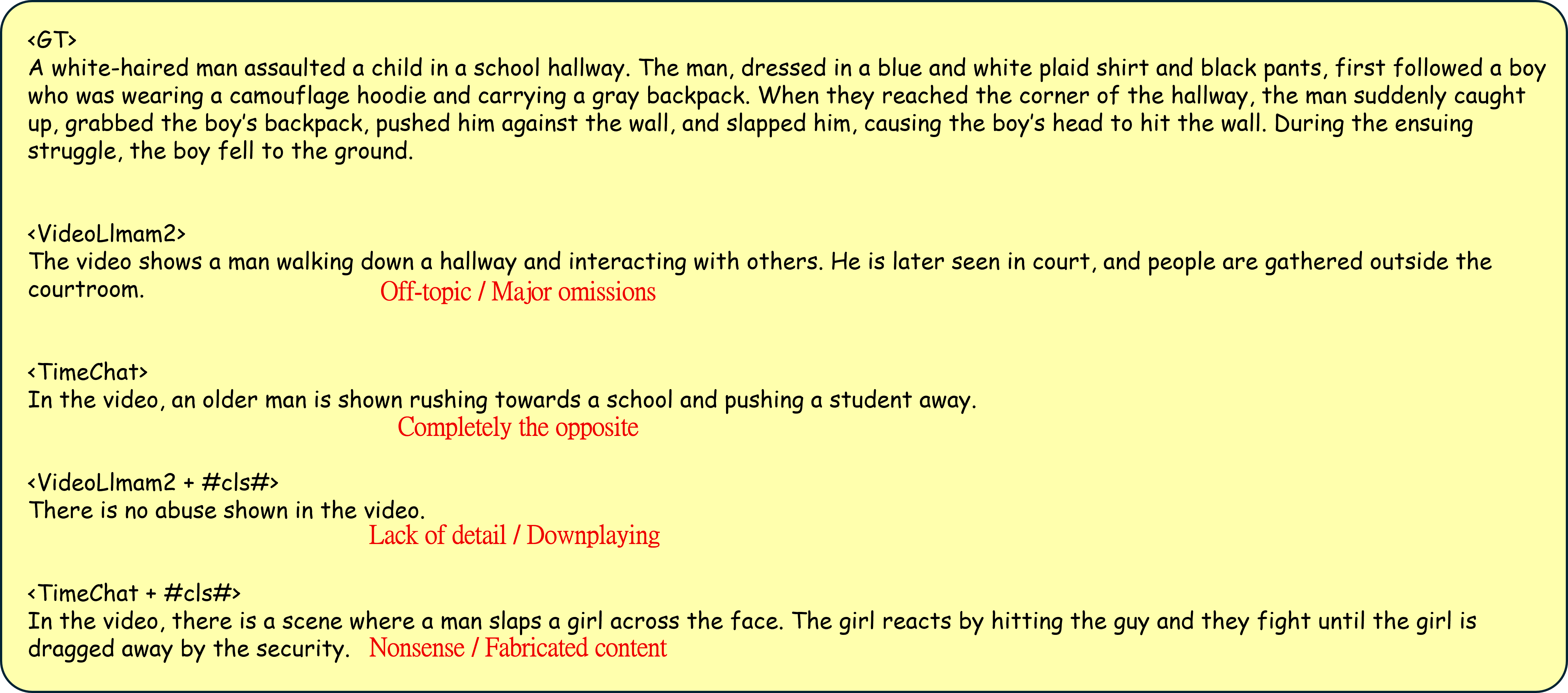}
	\caption{Directly applying current advanced video understanding and grounding models to anomalous videos often results in completely incorrect outputs, as the general sampling methods currently in use are unable to capture sufficient anomalous cues.}
	\label{fig:sup5}
\end{figure*}

Regarding our repeatedly emphasized point that "current VQA/VTG models perform poorly on open-ended questions," we present a concrete case in Fig.~\ref{fig:sup5}. As shown, when the model is confronted with open-ended and ambiguous questions such as "Please describe the abnormal event in the video, below is a list of possible anomalies:{XX, ...}" or "What anomaly occurred in the video, below is a list of possible anomalies:{XX, ...}", its performance remains quite limited even when provided with a list of possible anomalies. Furthermore, since most current multimodal large models adopt strategies such as segment sampling when handling video tasks, the model still struggles to accurately understand and describe the specific details of anomalous events, even when the category of the anomaly in the video is explicitly given.

We would like to emphasize that this situation does not stem from any inherent shortcomings in the performance of these general-purpose models, whose outstanding contributions to the field we fully acknowledge. The root cause of detection failures lies in the conflict between the focus on holistic cues in video understanding and the demand for fine-grained cues in the VAD domain. This is precisely the core motivation behind our proposed GtS framework and the VAGU benchmark.

\begin{table}[ht]
	\centering
	\caption{Performance Comparison of Different Sample Methods}
	\begin{tabular}{l c}
		\hline
		\textbf{Method} & \textbf{Score} \\
		\hline
		Qwen2.5-VL-7B + TC & 3.61 \\
		Qwen2.5-VL-7B + TC + Uniform Split & 4.02 \\
		Qwen2.5-VL-7B + TC + GtS & 5.50 \\
		\hline
	\end{tabular}
	\label{table:comparison}
\end{table}

To further verify that the performance improvement of the GtS framework does not simply result from the increased number of sampled frames due to video segmentation, we conducted an experiment on the Qwen-2.5-VL-7B model in which videos were uniformly divided into seven segments (the maximum number of segments observed for the GtS framework on the VAGU dataset), and a similar processing pipeline was employed for the video anomaly understanding task. The results are presented in Table.~\ref{table:comparison}. The experiments indicate that, although increasing the number of sampled frames provides more details, these additional non-anomalous details do not significantly enhance the model's ability to understand anomalies. Further analysis reveals that uniform segmentation often leads to fragmented events, which in turn exacerbates the model's hallucination problem.

\begin{figure*}[h]
	\centering
	\includegraphics[width=\linewidth]{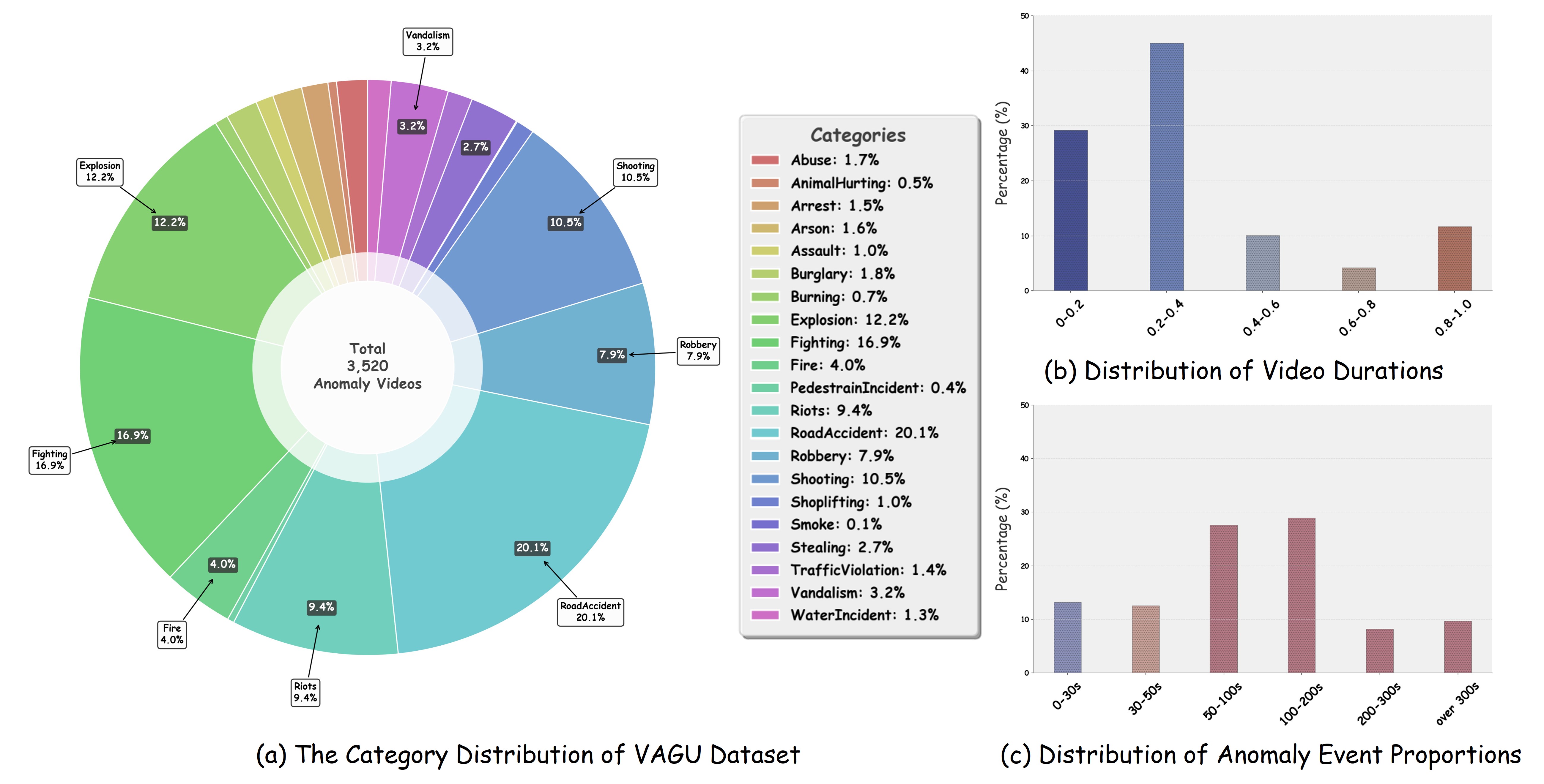}
	\caption{Presentation of statistical information regarding the proposed VAGU dataset.}
	\label{fig:benchmark}
\end{figure*}

\section{Dataset Detailed Information}
The VAGU dataset comprises 7,567 high-quality anomaly videos, each annotated with four labels for anomaly classification, understanding, grounding and QA.
The average video duration over 2,700 frames. This dataset covers four major domains—human criminal activities, natural disasters, traffic accidents, and animal-related injuries—encompassing 21 fine-grained categories. Fig.~\ref{fig:benchmark}(a) illustrates the distribution of specific categories, while Fig.~\ref{fig:benchmark}(b) and Fig.~\ref{fig:benchmark}(c) present statistical distributions of video duration and anomaly event proportion, respectively. Notably, certain categories (e.g., "Fire," "Arson," "Burning") exhibit semantic similarities; detailed definition criteria for differentiation can be found upside. Additionally, due to the limitations of existing VLMs, we encourage the use of multiple VLMs for ensemble-based completion of VAD tasks on the VAGU dataset.

\section{Detailed Comparison with Other Benchmarks}

To further elucidate the innovations and contributions of our proposed VAGU dataset, we conducted a comprehensive comparative analysis between VAGU and several leading existing benchmarks.

\textbf{CUVA}: CUVA introduced a QA dataset that encompasses both video understanding and video grounding tasks. However, as explicitly stated in Appendix A.4 of the CUVA paper, the grounding annotations in CUVA are automatically generated by large language models. This automated annotation approach is not only prone to labeling errors, but our experiments also revealed significant biases in the data. More importantly, CUVA primarily focuses on the semantic and causal understanding of anomalous events, and its experimental design and evaluation metrics do not assess grounding performance. The evaluation results for timestamps reported in Table 2 of the paper further confirm the inaccuracy of its grounding annotations. Additionally, CUVA does not provide a QA dataset.

\textbf{Holmes-VAU}: Holmes-VAU presented a large-scale video anomaly understanding benchmark dataset with hierarchical instructions. Although this dataset offers diverse annotations at different segment levels, the grounding of anomalous events also relies on large models to automatically merge uniformly segmented video captions. Similar to CUVA, this annotation method tends to yield insufficient grounding accuracy. Moreover, Holmes-VAU does not provide a QA dataset.

\textbf{VANE-Bench}: VANE-Bench, like CUVA, does not include evaluations related to anomaly grounding, but instead serves as a video QA dataset. Its evaluation strategy is limited to a binary scoring system: one point for each correct answer and zero for each incorrect answer. The dataset lacks annotations for both anomaly video understanding and anomaly video grounding, and only offers 559 QA pairs, which is substantially fewer than the 20,000+ QA pairs provided in our dataset.

\begin{table*}[ht]
	\centering
	\caption{Comparison of VAGU with Existing Benchmarks.}
	\begin{tabular}{lcccccc}
		\toprule
		\textbf{Dataset}   & \textbf{\#Videos} & \textbf{\#Anomaly Categories} & \textbf{QA Provided} & \textbf{\#QA Pairs} & \textbf{AU Provided} & \textbf{AG Provided} \\
		\midrule
		CUVA        &1000 &11   &     &       & Human   & VLM   \\
		Holmes-VAU  &5443 &15   &     &       & Human   & VLM   \\
		VANE-Bench  &325  &19   & \checkmark   & 559    &     &    \\
		VAGU (Ours) &7567 &21   & \checkmark   & 20,000+& Human   & Human  \\
		\bottomrule
	\end{tabular}
	\label{tab:benchmark_comparison}
\end{table*}

As shown in Table.~\ref{tab:benchmark_comparison}, the VAGU dataset not only surpasses existing benchmarks in annotation quality and task coverage, but also offers substantial improvements in the number of QA pairs, as well as in the granularity of anomaly understanding and grounding annotations. Our dataset provides a more reliable and comprehensive benchmark for the evaluation and research of multimodal large models on video anomaly analysis tasks.

\section{Detailed Taxonomy of Anomaly Categories in VAGU Dataset}
In this section, we will provide a systematic elucidation of anomaly categories in the VAGU dataset, with specialized emphasis on disambiguating semantically overlapping classes. The formal definitions and critical distinctions are structured as follows:

Fire-Related Incidents:
\begin{itemize}
	\item[$\bullet$]Fire: Non-intentional fires caused by factors such as electrical short circuits, kitchen grease fires, etc.
	\item[$\bullet$]Arson: The criminal act of intentionally igniting fires, often motivated by insurance fraud or revenge.
	\item[$\bullet$]Burning: A sustained combustion state (e.g., garbage incineration) or physiological damage from heat (e.g., skin burns).
	\item[$\bullet$]Smoke: Refers to either smoking behavior or smoke produced by fires/combustion (context-dependent).
\end{itemize}

Theft-Related Incidents:
\begin{itemize}
	\item[$\bullet$]Stealing: General theft without violence, including pickpocketing or opportunistic theft.
	\item[$\bullet$]Burglary: Illegal entry into premises (e.g., homes/stores) to commit theft, often involving property damage.
	\item[$\bullet$]Robbery: Violent theft involving force/threats to seize property, potentially causing physical harm.
	\item[$\bullet$]Shoplifting: Concealing merchandise to avoid payment in retail settings, often involving anti-theft countermeasures.
\end{itemize}

Violence-Related Incidents:
\begin{itemize}
	\item[$\bullet$]Fighting: Physical altercations between parties, potentially escalating from verbal disputes (no weapons specified).
	\item[$\bullet$]Assault: Unilateral violent acts (e.g., sudden attacks, object-throwing) with intent to harm.
	\item[$\bullet$]Abuse: Persistent physical/psychological harm (e.g., domestic/workplace violence, verbal degradation).
	\item[$\bullet$]Riots: Large-scale public disorder involving vandalism, looting, and arson by groups.
\end{itemize}

Traffic-Related Incidents:
\begin{itemize}
	\item[$\bullet$]Road Accident: Vehicle collisions/rollovers causing injuries, fatalities, or property damage.
	\item[$\bullet$]Traffic Violation: Breaches of traffic laws (e.g., illegal lane changes, speeding, red-light running).
	\item[$\bullet$]Pedestrian Incident: Accidents involving pedestrians (e.g., jaywalking collisions, falls).
\end{itemize}

Other Incidents
\begin{itemize}
	\item[$\bullet$]Explosion: Sudden energy releases from gas leaks, explosives, etc.
	\item[$\bullet$]Vandalism: Intentional property damage (e.g., graffiti, public facilities destruction).
	\item[$\bullet$]Water Incident: Aquatic hazards (e.g., drownings, boat collisions, unrecovered falls).
	\item[$\bullet$]Animal Hurting: Acts harming animals (e.g., pet abuse, wildlife poaching).
	\item[$\bullet$]Shooting: Firearm discharges causing injuries or property damage.
	\item[$\bullet$]Arrest: Law enforcement detaining suspects through legal procedures.
\end{itemize}

\section{Ethical Review Process}
This study strictly adheres to data ethics guidelines and implements full-process compliance management:

\subsection{Data Acquisition Compliance}
\begin{itemize}
	\item[$\bullet$]Strict compliance with copyright regulations of platforms such as YouTube, BiliBili, and TikTok.
	\item[$\bullet$]Adoption of a double-blind annotation mechanism; sensitive video frames containing faces, license plates, etc., are processed with mosaics or edited out.
\end{itemize}

\subsection{Content Screening Criteria}
\begin{itemize}
	\item[$\bullet$]Privacy Protection: Exclude content with identifiable biometric features or geolocation information.
	\item[$\bullet$]Copyright Compliance: Exclude materials containing unauthorized film clips or copyrighted music.
	\item[$\bullet$]Violence Control: Graphic violence scenes are removed or processed with mosaics.
	\item[$\bullet$]Child Protection: Delete footage involving minors in dangerous situations.
	\item[$\bullet$]Gender Equality:
	
	- Filter content with sexual implications or gender-discriminatory language.
	
	- Remove visual symbols objectifying specific gender groups.
	\item[$\bullet$]Illegal Content: Block content involving drug trafficking, weapon display, or other illegal activities.
\end{itemize}

\subsection{Quality Control System}
\begin{itemize}
	\item[$\bullet$]Each sample undergoes cross-validation by three certified ethical reviewers.
	\item[$\bullet$]Maintain comprehensive operation logs for the entire process.
\end{itemize}

\section{Case Study}

\begin{figure*}[h]
	\centering
	\includegraphics[width=0.85\linewidth]{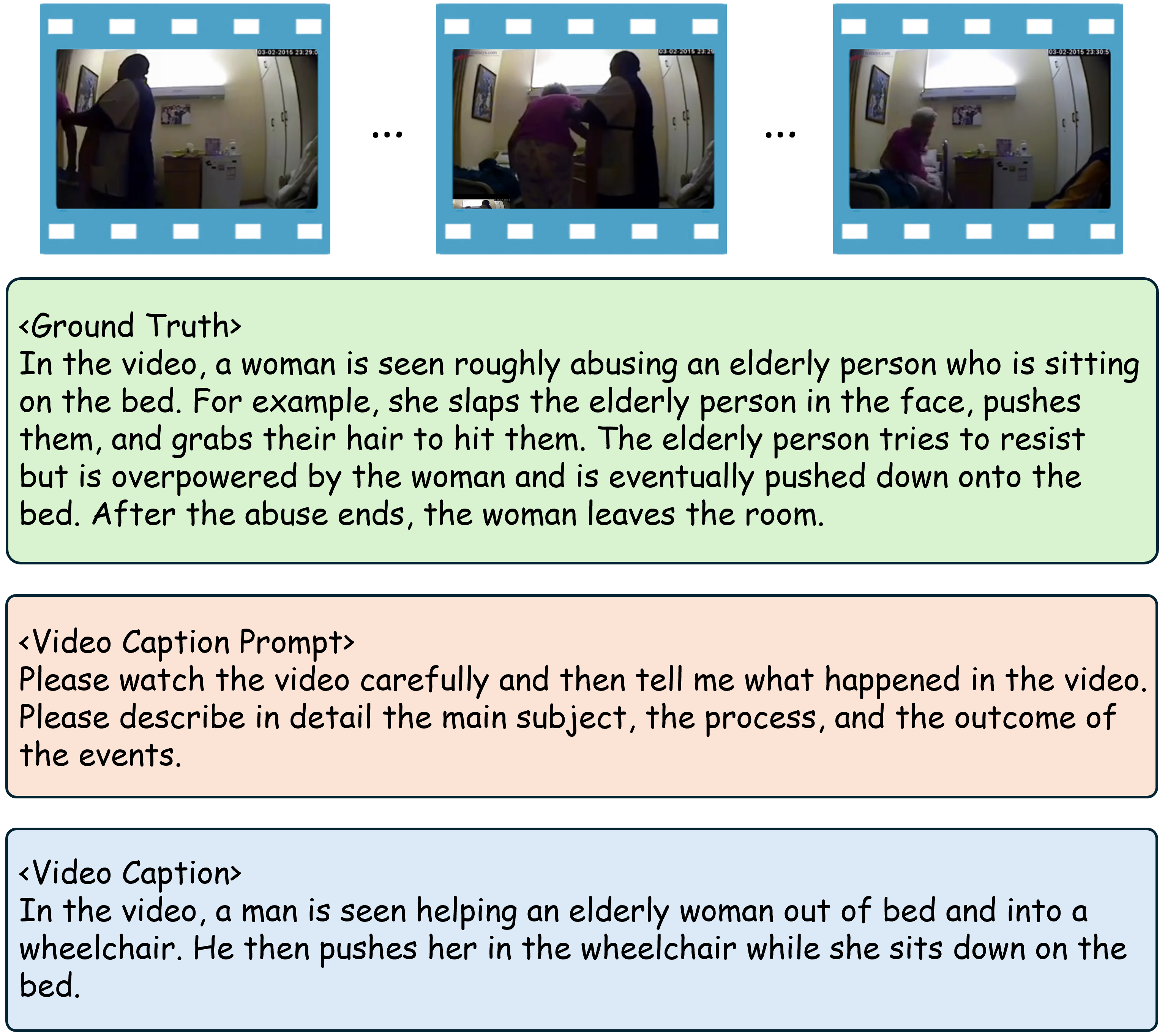}
	\caption{The first part of inference in the GtS framework involves directly using the VLM to generate captions for the video.}
	\label{fig:case 1}
\end{figure*}

\begin{figure*}[h]
	\centering
	\includegraphics[width=0.85\linewidth]{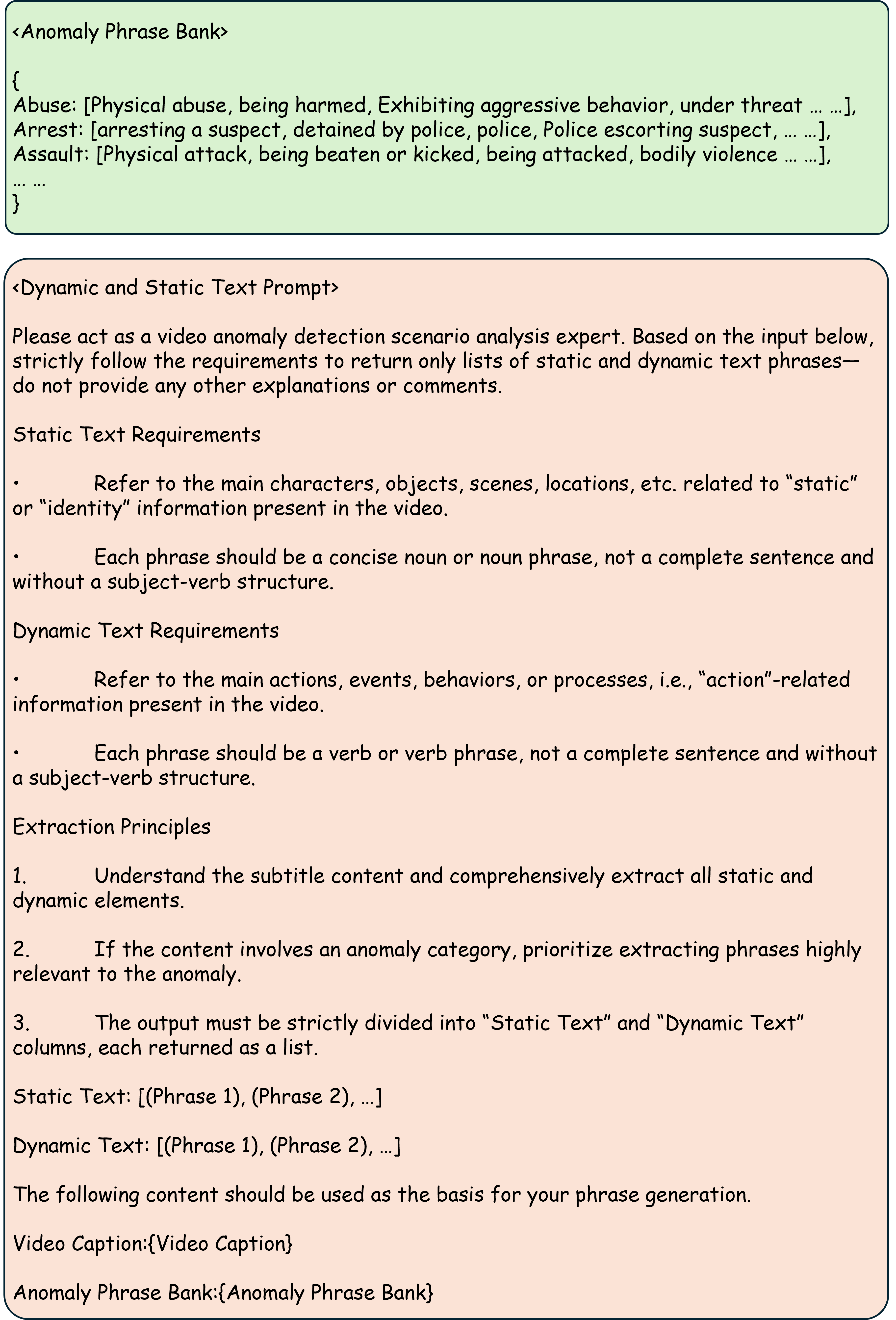}
	\caption{The second part of inference in the GtS framework utilizes a pre-generated phrase bank and the video caption to produce static and dynamic guiding texts.}
	\label{fig:case 2}
\end{figure*}

\begin{figure*}[h]
	\centering
	\includegraphics[width=0.85\linewidth]{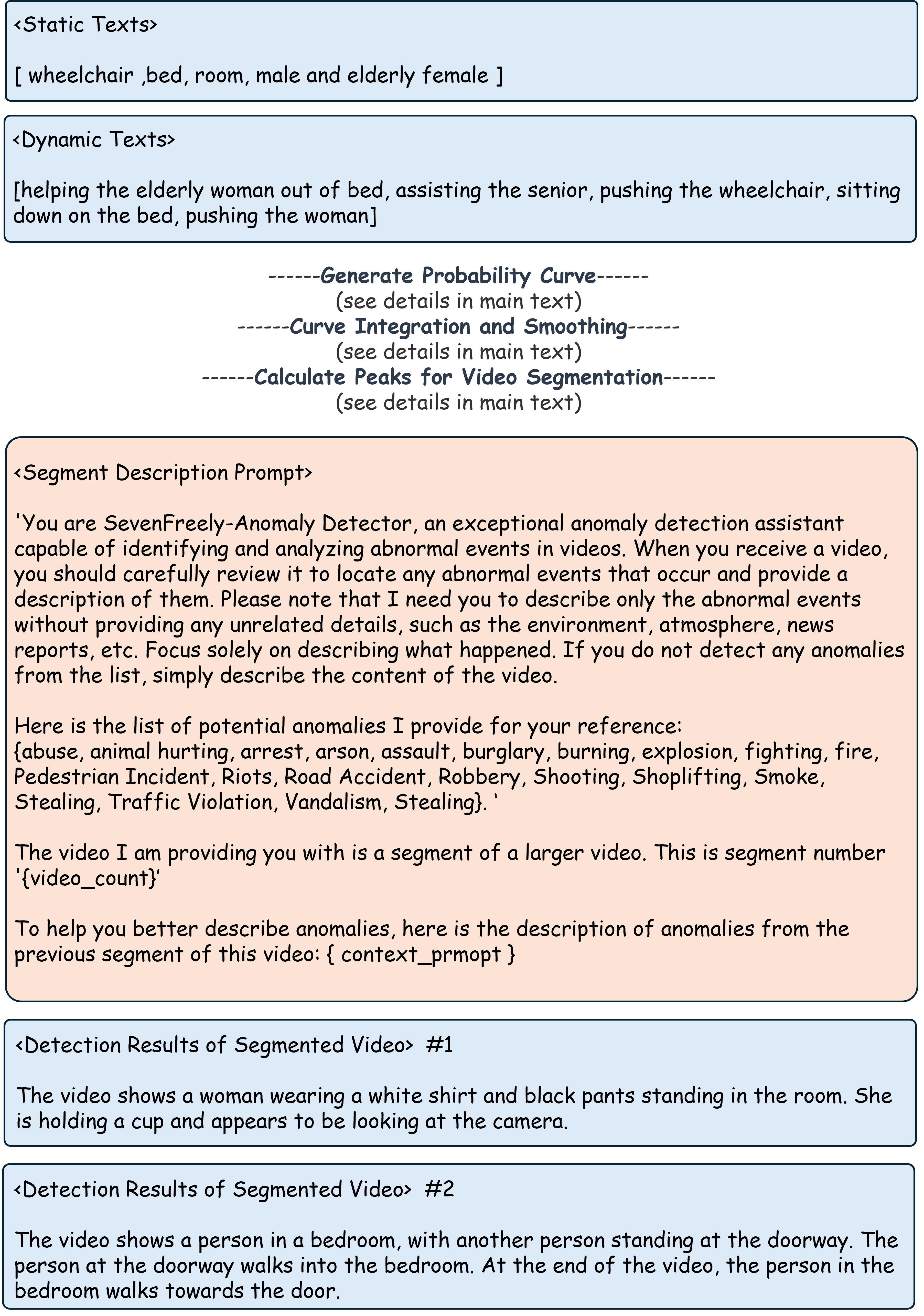}
	\caption{The third part of inference in the GtS framework applies the previously mentioned algorithm to segment the entire video, and then performs anomaly detection on the segmented video clips.}
	\label{fig:case 3}
\end{figure*}

\begin{figure*}[h]
	\centering
	\includegraphics[width=0.85\linewidth]{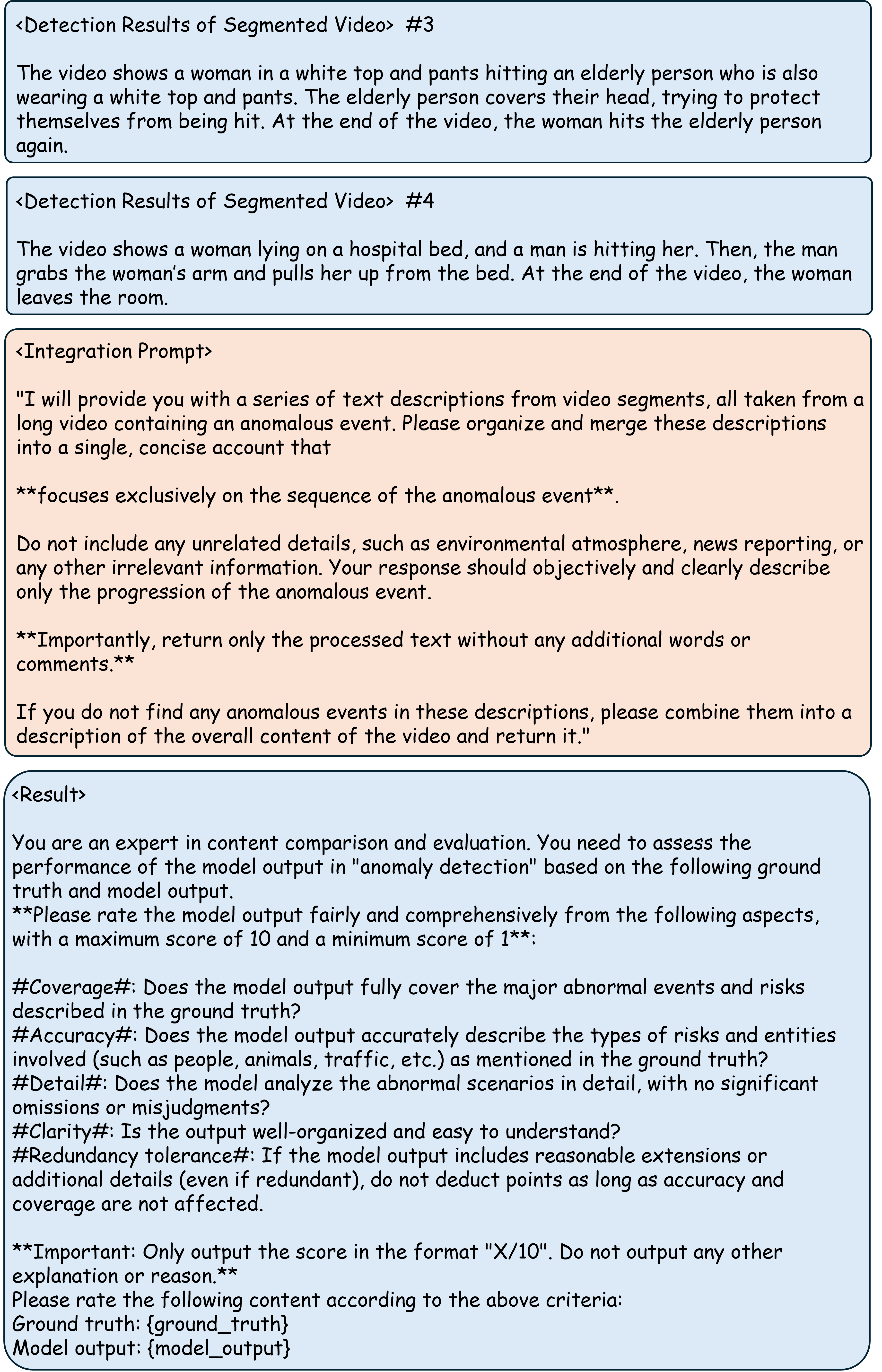}
	\caption{The fourth part of inference in the GtS framework merges the detection results of all segmented video clips and uses an LLM to score them.}
	\label{fig:case 4}
\end{figure*}

In this section, we provide a detailed exposition of an inference process based on the GtS framework in conjunction with prompts. The specific procedure is illustrated in Fig.~\ref{fig:case 1}, Fig.~\ref{fig:case 2}, Fig.~\ref{fig:case 3}, Fig.~\ref{fig:case 4}. The GtS framework score is $8/10$, and the VQA model score is $2/10$.

\end{document}